\documentclass[10pt,twocolumn]{article}

\usepackage[margin=0.75in,columnsep=0.25in]{geometry}
\usepackage[numbers,sort&compress]{natbib}
\usepackage{authblk}

\usepackage{algorithm}
\usepackage{algorithmic}
\usepackage{bm}
\usepackage[caption=false]{subfig}
\usepackage{multirow}
\usepackage{graphicx}
\usepackage{amsmath,amssymb}
\usepackage{booktabs}
\usepackage{placeins}
\usepackage{caption}
\usepackage[colorlinks,
            linkcolor=blue,
            anchorcolor=blue,
            citecolor=blue]{hyperref}

\title{Brain foundation model-guided source-selective domain adaptation for cross-subject EEG decoding}

\author[1,4]{Jinzhou Wu}
\author[1]{Baoping Tang\thanks{Corresponding author. E-mail: \href{mailto:bptang@cqu.edu.cn}{bptang@cqu.edu.cn}}}
\author[1]{Qikang Li}
\author[1]{Yi Wang}
\author[1]{Cheng Li}
\author[2,3]{Shujian Yu}

\affil[1]{State Key Laboratory of Mechanical Transmission for Advanced Equipment, College of Mechanical and Vehicle Engineering, Chongqing University, Chongqing 400044, China}
\affil[2]{Department of Computer Science, Vrije Universiteit Amsterdam, Amsterdam 1081 HV, The Netherlands}
\affil[3]{Machine Learning Group, UiT--The Arctic University of Norway, Tromso 9019, Norway}
\affil[4]{College of Computing and Data Science, Nanyang Technological University, Singapore 639798, Singapore}

\date{}

\begin{document}

\maketitle

{\renewcommand{\thefootnote}{}\footnotetext{Code is available at: \url{https://github.com/wjz114/BFM-MSDA}.}}

\begin{abstract}
Cross-subject motor-imagery electroencephalography (MI-EEG) decoding remains challenging because substantial inter-subject variability can cause both negative transfer from poorly matched source subjects and persistent distribution discrepancies between source and target domains. Existing multi-source domain adaptation methods often incorporate all available source domains or estimate source relevance using signal-level or task-specific representations, while distribution alignment is frequently performed only at the feature level. These limitations may introduce irrelevant source knowledge and fail to preserve class-discriminative structures across subjects. In this study, we propose a brain foundation model-guided multi-source domain adaptation framework (BFM-MSDA) for cross-subject MI-EEG decoding. The method first utilizes representations learned by a pretrained brain foundation model to estimate source--target compatibility and retrieve target-relevant sources. Subsequently, a relevance-weighted dual alignment strategy is applied to the selected sources and the unlabeled target domain. Specifically, Cauchy--Schwarz (CS) divergence is used to reduce discrepancies in marginal feature distributions, while conditional Cauchy--Schwarz (CCS) divergence further aligns class-dependent decision distributions. Source relevance is incorporated into both alignment terms so that more transferable source subjects contribute more strongly to adaptation. Experiments on two public MI-EEG benchmarks achieve average accuracies of 86.16\% and 78.41\%, outperforming representative cross-subject decoding and domain adaptation methods. Additional experiments with a large source pool further demonstrate that target-aware source retrieval improves scalability while mitigating negative transfer. These results highlights the importance of informed source selection under cross-subject distribution shift.
\end{abstract}

\noindent\textbf{Keywords:} Unsupervised domain adaptation; Brain-computer interface; Motor imagery; EEG foundation model; Source selection

\section{Introduction}

Motor-imagery electroencephalography (MI-EEG) decoding is an important paradigm in brain--computer interfaces (BCIs), with applications in assistive control and neurorehabilitation \cite{razaDeepLearningBased2020,chen_integration_2025,hortalSVMbasedBrainMachine2015}. Deep neural networks have substantially advanced MI-EEG decoding by learning discriminative spatio-temporal representations directly from EEG signals \cite{lawhernEEGNetCompactConvolutional2018,songEEGConformerConvolutional2023,CHEN2026131858}. However, their performance often degrades when applied to unseen subjects because EEG distributions vary considerably across individuals owing to differences in neurophysiology, mental strategy, and recording conditions \cite{yinGITGANGenerativeIntersubject2024,liCrosssubjectGeneralizationEEG2026,LIANG2025130875}. Consequently, a model trained on existing subjects generally requires additional calibration before it can be reliably deployed for a new user, limiting the practical scalability of EEG decoding systems \cite{DU2026134774}.

Domain adaptation (DA) provides a natural framework for this challenge. DA aims to exploit knowledge from labeled source domains to improve learning in a target domain with few or no target labels \cite{wuTransferLearningEEGBased2022, guoCrosssessionNonstationaryAttentionbased2025}, which is a common practical BCI setting. Early DA methods primarily focus on aligning the marginal distributions of learned feature representations between source and target domains, commonly achieved either explicitly through distance metric-based measures such as Maximum Mean Discrepancy (MMD) \cite{grettonKernelTwosampleTest2012} or implicitly via adversarial training \cite{longDeepTransferLearning2017}. However, marginal alignment alone does not guarantee preservation of class-discriminative structure in the target domain. In MI-EEG decoding, different subjects may have similar global feature distributions while exhibiting different decision boundaries. Therefore, subsequent efforts have sought to incorporate implicit alignment of conditional distributions or joint distributions to preserve class discriminability across domains \cite{weiMultiSourceTransferJoint2023,zhangDiscriminativeJointProbability2020}. For example, Hong et al. \cite{hongDynamicJointDomain2021} proposed a dynamic adversarial network (DJDAN) and designed a local discriminator to align the conditional distribution of the classifier's predictions.

Recent studies suggest that using diverse data from multiple source domains can improve model robustness and generalization, leading to the development of multi-source domain adaptation (MSDA) approaches \cite{ZHAO2026134158}. For instance, Liu et al. \cite{liuMultiSourceTransferLearning2023} proposed a domain-adversarial neural network (DANN) in which the final prediction is obtained by weighting the outputs of all source models. Although access to multiple sources provides a richer knowledge repository, it also introduces substantial heterogeneity across sources. Not every source subject is equally transferable to a given target, and indiscriminately combining poorly matched subjects may cause negative transfer and increase the computational cost of adaptation as the source repository grows \cite{yinGITGANGenerativeIntersubject2024,LIANG2025130875}. Therefore, effective cross-subject MSDA requires suppressing negative transfer from irrelevant or poorly matched source subjects. To mitigate this issue, some studies have explored strategies for source selection. For example, Adaptive Source Joint Domain Adaptation (ASJDA) \cite{liuEnhancingEEGBasedCrossSubject2021} uses Jensen--Shannon divergence between EEG distributions to filter source subjects, while recent approaches have explored source similarity in structured feature spaces to suppress negative transfer \cite{dingRSSDARiemannianSource2026}. However, raw EEG distributions are highly sensitive to noise and inter-subject variability, whereas representations learned with a downstream adaptation model may themselves be affected by the source domains used for training. These limitations highlight the need for an MSDA framework that can reliably assess whether knowledge from one subject will transfer effectively to another.

Brain foundation models (BFM) offer a new opportunity for subject-robust EEG adaptation \cite{dengBrainFoundationModels2026}. By pretraining on large and heterogeneous neural recordings, these models can provide latent representations that encode reusable structure across subjects, tasks, and acquisition conditions \cite{zhouBrainFoundationModels2025}. The BFM latent embedding is expected to implicitly capture representational variance in EEG signals and reflect such neurological relevance at a latent level \cite{liaoEEGTuneDataefficientFinetuning2026}. Due to this property, existing studies have primarily treated these models as feature extractors or pretrained backbones that are subsequently fine-tuned for downstream prediction, such as motor imagery decoding \cite{jiang_large_2024}, emotion recognition \cite{wangCBraModCrissCrossBrain2025}, sleep staging \cite{yuanBrainWaveBrainSignal2024}, and mental disorder diagnosis \cite{dengBrainFoundationModels2026}. However, their utility for DA remains largely unexplored. As BFMs may provide an external representation space for assessing compatibility between two candidate subjects, this motivates the use of pretrained brain representations as a transferability prior for source retrieval in EEG MSDA.

Motivated by this opportunity, we propose a brain foundation model-guided multi-source domain adaptation (BFM-MSDA) framework for subject-independent MI-EEG decoding. The framework first uses a pretrained BFM to encode the source and target subjects and retrieve the source subjects most relevant to the current target. It then performs dual-level alignment over the selected sources, using Cauchy-Schwarz (CS) divergence for feature-level alignment and conditional Cauchy-Schwarz (CCS) divergence for decision-level alignment. This design explicitly addresses both representation shift and decision-level mismatch, while relevance-weighted source aggregation reduces negative transfer and unnecessary computation. To our knowledge, this is the first endeavor to integrate the representational power of large BFMs with CS divergence-based alignment for cross-subject MI-EEG decoding.

The key contributions of this work are as follows:
\begin{enumerate}
\item We introduce a BFM-guided source retrieval strategy that uses pretrained brain representations as a transferability prior for cross-subject adaptation, reducing negative transfer and computational burden compared to adapting across all source domains.
\item We develop a relevance-weighted dual-alignment framework based on CS and conditional CS divergences, which jointly aligns feature distributions and decision distributions across selected sources and the target.
\item Extensive experiments on two benchmark datasets demonstrate superior cross-subject decoding accuracy against representative and state-of-the-art baselines. Additional experiments with larger source pools validated the scalability of the proposed source selection strategy, which substantially reduces computational cost while maintaining strong performance.
\end{enumerate}

\section{Related Works}
\subsection{EEG-based motor imagery classification}
The advancement of machine learning has profoundly shaped the development of BCI systems. Innovations in CNNs have driven substantial progress in EEG decoding. Foundational works by Schirrmeister et al. \cite{schirrmeisterDeepLearningConvolutional2017} established deep ConvNet and Shallow ConvNet architectures with temporal-spatial filtering layers optimized for EEG's multidimensional characteristics. The subsequent EEGNet \cite{lawhernEEGNetCompactConvolutional2018} introduced parameter-efficient depthwise convolutions while maintaining interpretability. Building on these principles, recent studies have enhanced model efficacy through multiple strategies. Chen et al. \cite{chenMultiattentionAdaptationNetwork2022} addressed the location of critical EEG channels and designed an end-to-end channel selection strategy for MI recognition. Tao et al. \cite{taoADFCNNAttentionBasedDualScale2024} designed parallel CNN branches to extract multi-scale features and fuse them with a self-attention module. More recently, Chen et al. \cite{CHEN2026131858} proposed DBTSP-Net, a dual-branch temporal--spatial parallel network with adaptive feature fusion and automated hyperparameter optimization for subject-specific MI-EEG decoding. An adaptive frequency-aware Transformer (AFFormer) \cite{CHANG2026111144} further introduced subject-adaptive spectral decomposition followed by Transformer-based spatiotemporal modeling. These developments highlight an architectural evolution toward combining localized feature extraction with multi-level context modeling, significantly advancing MI classification performance. Despite these advancements, vanilla deep learning approaches often perform poorly when trained on data from one subject and tested on another, necessitating time-consuming calibration procedures for each new user \cite{liCrosssubjectGeneralizationEEG2026}. This limitation has motivated the development of DA techniques specifically designed for cross-subject EEG classification.

\subsection{Domain adaptation in motor imagery classification}
DA has become a powerful strategy to overcome the limitations of cross-subject EEG classification, enabling knowledge transfer from labeled source subjects to target subjects with no labeled data \cite{liCrosssubjectGeneralizationEEG2026}. Early approaches to cross-subject EEG classification focused on projecting EEG data into a shared feature space to align marginal probability distributions. For example, Euclidean Alignment (EA) \cite{heTransferLearningBrain2020} has been proposed to directly align EEG trials from different subjects by transforming them into Euclidean space. Hong et al. \cite{hongDynamicJointDomain2021} proposed DJDAN to dynamically balance marginal and conditional distribution alignment through adversarial learning, whereas Song et al. \cite{songGlobalAdaptiveTransformer2023} combined Transformer-based global dependency modeling with adversarial domain adaptation. Liu et al. \cite{liuMultiSourceTransferLearning2023} addressed the challenge of integrating multiple source domains by developing a unified multi-source optimization framework that combines weighted predictions from multiple domains to produce the final classification result. More recently, Liang et al. \cite{LIANG2025130875} proposed MVC-former adaptation, which extracts complementary frequency-specific spatiotemporal representations using parallel convolution and Transformer modules for cross-subject MI classification. In MSDA, not all source subjects contribute equally to adaptation performance. Using all source subjects indiscriminately may lead to negative transfer and unnecessary computational costs during adaptation \cite{yinGITGANGenerativeIntersubject2024}. To address this issue, ASJDA \cite{liuEnhancingEEGBasedCrossSubject2021} filters source subjects for adaptation based on their JS divergence from the target. Zhao et al. \cite{ZHAO2026134158} proposed a curriculum-learning-based MSDA framework for cross-subject EEG emotion recognition, in which source domains are clustered and a multi-armed bandit strategy selects target samples during multi-order moment matching. These studies demonstrate that both source-domain relevance and the adaptation process itself can critically affect multi-source transfer performance.

\subsection{Brain foundation model}
The emergence of large Brain Foundation Models (BFMs) has revolutionized EEG analysis through large-scale self-supervised pretraining. Inspired by the success of large language models (LLMs), BFMs are pre-trained on large-scale physiological datasets to learn robust, generalizable neural representations \cite{zhouBrainFoundationModels2025}. The pre-training datasets are typically heterogeneous, encompassing a wide variety of tasks such as motor imagery classification \cite{chenUniNTFMUnifiedFoundation2026}, disease diagnosis \cite{dengBrainFoundationModels2026}, emotion recognition \cite{wangCBraModCrissCrossBrain2025}, and other neurophysiological tasks \cite{wangCBraModCrissCrossBrain2025}. The diversity in the training corpus enables BFMs to learn a universal feature space that generalizes across tasks and populations. To enhance generalizability across BCI tasks, Jiang et al. presented LaBraM \cite{jiang_large_2024}, a large-scale foundation model that segments EEG into channel patches and employs vector-quantized neural spectrum prediction for pretraining. Moreover, BrainWave \cite{yuanBrainWaveBrainSignal2024}, trained on more than 40,000 hours of invasive and non-invasive brain recordings, was proposed to achieve superior performance for diagnosing neurological disorders. Wang et al. proposed CBraMod \cite{wangCBraModCrissCrossBrain2025}, a criss-cross transformer that separately models spatial-temporal EEG dependencies and the comprehensive structural characteristics of EEG signals. 
The BFM latent embedding is expected to implicitly capture the intersubject variance and reflect such neurological relevance at a latent level.

While BFMs have shown strong potential in learning universal EEG representations, their role in subject-independent DA remains insufficiently explored. Our study bridges this gap by proposing a novel framework that harnesses BFM-derived embeddings to quantify inter-subject similarity, thereby enabling effective source selection and distribution alignment.

\section{Methodology}
This section presents the problem setup, notation, and the methodology of the proposed BFM-MSDA framework for subject-independent MI decoding tasks. Fig.~\ref{fig_framework} illustrates the BFM-MSDA framework and its key components.

\subsection{Problem definition and notation}
We study subject-independent MI decoding in an unsupervised DA setting, where labeled prior subjects are available, but the incoming target subject has no labels during training. Given a repository of labeled source subjects and an unlabeled target subject, our goal is to retrieve a compact set of source subjects most compatible with the target and adapt them without using target labels.

Formally, let \(\{D_s\}_{s=1}^S\) denote the labeled EEG datasets from a total of \(S\) source subjects, where the $s$-th subject's dataset is given by \(D_s = \{(\bm{x}^{s}_{i}, y^{s}_{i})\}_{i=1}^{M_s}\). Here, \(\bm{x}^{s}_{i} \in \mathbb{R}^{C \times T}\) represents an EEG trial with \(C\) channels and \(T\) time points, and \(y^{s}_{i} \in \{1, \dots, K\}\) is the class label corresponding to one of \(K\) MI tasks. The unlabeled EEG data from a target subject is denoted as \(D_{t}=\{(\bm{x}^{t}_{j})\}^{M_t}_{j=1}\).

The goal is to learn a model \(\mathcal{M}=g\circ f\), where the feature extractor \(f\colon\mathbb{R}^{C\times T}\to\mathbb{R}^{d}\) maps raw MI-EEG signals to a latent representation with dimension $d$, and the classifier \(g\colon\mathbb{R}^{d}\to\{1,\dots, K\}\) outputs predicted class for MI tasks, such that \(\mathcal{M}\) generalizes effectively to \(D_{t}\) in an unsupervised DA setting.

\begin{figure*}[!t]
\centering
\includegraphics[width=0.92\textwidth]{Overall_framework.png}
\caption{Overview of the proposed BFM-MSDA framework for subject-independent MI decoding: (a) source selection with a pretrained BFM, where trial-level embeddings \(\mathbf{h}_1,\ldots,\mathbf{h}_S\) and \(\mathbf{h}_T\) yield pairwise CS distances \(d_{S_i,T}\) used to retain target-relevant sources; (b) MSDA with feature-level alignment (FLA) via weighted CS on \(p(z)\) and decision-level alignment (DLA) via conditional CS on \(p(y|z)\) over selected source--target and source--source pairs; (c) testing, where the trained feature extractor and classifier predict MI labels.}
\label{fig_framework}
\end{figure*}

\subsection{Preliminary knowledge on Cauchy-Schwarz divergence}
Distributional alignment remains challenging due to the high dimensionality of EEG signals and the continuous nature of the latent representations. Many previous DA methods focus on aligning class-conditional feature distributions \(p(\bm{x}|y)\) rather than posterior label distributions \(p(y|\bm{x})\), due to relative simplicity of estimating \(p(\bm{x}|y)\) with $y$ being a discrete variable. However, from a Bayesian standpoint, the joint distribution can be factorized as \(p(\bm{x},y)=p(\bm{x})p(y|\bm{x})\), indicating that aligning the joint distribution effectively requires separate alignment of both marginal feature distribution and conditional label distribution. Aligning \(p(\bm{x}|y)\) alone does not guarantee alignment of \(p(y|\bm{x})\), which directly relates to decision boundaries critical for MI decoding tasks.

The CS divergence derives from the Cauchy-Schwarz inequality for square-integrable functions \(p(\bm{x})\) and \(q(\bm{x})\), which states:
\begin{equation}\label{eq:cs_inequal}
\left[ \int p(\bm{x})q(\bm{x})d\bm{x} \right]^2 \leq \int p^2(\bm{x})d\bm{x} \int q^2(\bm{x})d\bm{x},
\end{equation}
with equality holds only if \(p(\bm{x})\) and \(q(\bm{x})\) are linearly dependent. The CS divergence quantifies the discrepancy between \(p(\bm{x})\) and \(q(\bm{x})\) by taking the logarithm of the ratio between the left-hand side and the right-hand side of Eq.~\ref{eq:cs_inequal}:
\begin{equation}\label{eq:cs}
\begin{aligned}
D_{\mathrm{CS}}(p\|q) &=-\log \left(\frac{\left|\int p(\mathbf{x}) q(\mathbf{x}) d \mathbf{x}\right|^2}{\int|p(\mathbf{x})|^2 d \mathbf{x} \int|q(\mathbf{x})|^2 d \mathbf{x}}\right) \\
& =-2 \log \left(\int p(\mathbf{x}) q(\mathbf{x}) d \mathbf{x}\right) \\
&\quad +\log \left(\int p(\mathbf{x})^2 d \mathbf{x}\right) +\log \left(\int q(\mathbf{x})^2 d \mathbf{x}\right).
\end{aligned}
\end{equation}

Obviously, \(D_{\text{CS}}(p\|q) \geq 0\) with equality if \(p(\bm{x}) = q(\bm{x})\).

The CS divergence enables straightforward estimation for \(p(\bm{x})\) and \(q(\bm{x})\) without distributional assumptions. It offers several advantages over other distance measures in DA. It is symmetric, admits closed-form solutions for mixture-of-Gaussians (MoG) \cite{kampaClosedformCauchyschwarzPDF2011}, and provides provably tighter generalization error bounds compared to Kullback-Leibler (KL) divergence \cite{yin_domain_2024}. Moreover, CS divergence avoids the logarithmic singularity of KL divergence \(D_{KL}(P \parallel Q)=\int p(x)\log\frac{p(x)}{q(x)}\,dx\), which diverges when \(q(x)\) approaches zero \cite{yin_domain_2024}.

The conditional CS (CCS) divergence \cite{yuConditionalCauchySchwarzDivergence2024} extends the CS divergence from Eq.~\ref{eq:cs} to conditional distributions, which measures the discrepancy between two conditional distributions \(p(y|\bm{x})\) and \(q(y|\bm{x})\), even though \(x\) is a high-dimensional continuous random variables, which makes it well suited for DA tasks:
\begin{equation}\label{eq:ccs}
\begin{aligned}
D_{\mathrm{CCS}}(p\|q)
&:= -2\log \left( \iint p(y|\bm{x})q(y|\bm{x})\,d\bm{x}\,dy \right) \\
&\quad + \log \left( \iint p^2(y|\bm{x})\,d\bm{x}\,dy \right) \\
&\quad + \log \left( \iint q^2(y|\bm{x})\,d\bm{x}\,dy \right),
\end{aligned}
\end{equation}

In our scenario, \(x\) represents input EEG data or latent representations, and \(y\) denotes classification logits or probabilistic labels. Details on estimating CS and CCS divergences from a mini-batch of samples are provided in~\ref{app:ccs}.

\subsection{Target-aware source selection with a pretrained BFM}
\label{subsec:source_selection}
Indiscriminate use of all available source subjects during MSDA introduces negative transfer and excessive computational burden. To address this, our framework selects source subjects whose EEG data distributions are similar to that of the target subject with the guidance of BFM.

\subsubsection{Feature extraction and aggregation with LaBraM}
LaBraM is a unified foundation model pretrained on roughly 2,500 hours of heterogeneous EEG data spanning 20 datasets. During pretraining, LaBraM reconstructs original neural codes from masked EEG channel patches, thereby enhancing its ability to capture meaningful patterns in brain activity data. Notably, the datasets used in our study were not included in LaBraM's pretraining \cite{jiang_large_2024}, thereby eliminating the risk of data leakage.

The process of extracting subject-level EEG representations using LaBraM is illustrated in the bottom panel of Fig.~\ref{fig_framework}. Initially, EEG recordings are segmented into fixed-length, non-overlapping temporal patches, each comprising 200 samples, corresponding to 1 second of EEG activity at a sampling rate of 200 Hz. For each patch, LaBraM extracts a latent feature vector that incorporates temporal and spatial embeddings. Trial-level embeddings are aggregated into subject-level representations, allowing each subject to be summarized as a latent distribution for subsequent similarity estimation and source retrieval. For subject with \(M\) trials, we obtain a set of trial-level representations \(\{\mathbf{h}_i\}_{i=1}^{M}\), where each \(\mathbf{h}_i \in \mathbb{R}^{200}\). These trial-level feature vectors are then used to estimate the distribution \(p(\mathbf{h})\) for subsequent CS divergence computation in source selection.

\subsubsection{Source selection via inter-Subject similarity}
The discrepancies between source and target subjects are quantified by computing the CS divergence between their respective feature distributions:
\begin{equation} \label{eq:pairdist}
    d_{s,t} = D_{\mathrm{CS}}\left(p(\mathbf{h}_s) \,\|\, p(\mathbf{h}_t)\right), \quad s = 1,\dots,S,
\end{equation}
where $S$ denotes the total number of candidate source subjects, \(p(\mathbf{h}_s)\) and \(p(\mathbf{h}_t)\) represent the empirical distribution of trial-level feature vectors extracted by the pre-trained LaBraM-Base (5.8M parameters) model for source $s$ and target $t$.

To ensure effective adaptation, we adopt an adaptive soft-gating strategy to select a subset of close sources from the $S$ candidates. We convert source-target distances into normalized relevance weights and retain only the sources whose relevance exceeds a uniform-prior baseline. Specifically, for each candidate source $s$, we define a normalized relevance weight \(\tilde{w}_{s,t}\) by
\begin{equation} \label{eq:relevanceweight}
\tilde{w}_{s,t} = \frac{\exp(-d_{s,t}/\tau_t)}{\sum_{s'} \exp(-d_{s',t}/\tau_t)},
\end{equation}
where the temperature parameter \(\tau_t\) is self-calibrated for target subject $t$ as the median of the distances \(\{d_{s,t}\}_{s=1}^S\).

The subset is then formed by retaining sources whose weights satisfy \(\tilde{w}_{s,t}>1/S\), where \(1/S\) denotes the uniform prior over candidate sources. If this rule yields an empty set, the top-3 sources with the largest \(\tilde{w}_{s,t}\) are selected. The resulting subset contains $N$ sources for target $t$, where $N$ is determined adaptively and may vary across target subjects. In this way, downstream adaptation relies only on source subjects that are representation-wise most compatible with the target.

\subsection{Multi-source domain adaptation with CS divergences}
Let \(\bm{z}=f(\bm{x})\in\mathbb{R}^{d}\) be the latent feature extracted by the feature extractor of the backbone network. We propose aligning both feature-level and decision-level distributions between source and target domains using CS divergence, which learns domain-invariant representations while ensuring consistent decision boundaries.

\subsubsection{Feature-level alignment}
We first align the marginal distribution of learned features \(p(\mathbf{z})\). Euclidean alignment (EA) \cite{heTransferLearningBrain2020} is applied to reduce inter-subject distribution discrepancy by aligning the EEG trials from each subject to a common reference space. EA is performed by calculating the arithmetic mean of covariance matrices \textit{R} for the \textit{i}th EEG trial of a subject:
\begin{equation} \label{eq:EA}
\tilde{X}_i=\bar{R}^{-1 / 2} X_i,
\end{equation}
where \(\bar{R}=\frac{1}{M} \sum_{i=1}^M X_i X_i^T\) and \textit{M} represents the total number of EEG trials of the subject.

Different sources contribute unequally to adaptation. Therefore, we dynamically assign weights to the selected sources based on their CS divergence from the target:
\begin{equation} \label{eq:source_weights}
\omega_s = \frac{\exp\left(-D_{\text{CS}}(p_{s}(\bm{z}) \| p_{t}(\bm{z}))\right)}{\sum_{s'=1}^N \exp\left(-D_{\text{CS}}(p_{s'}(\bm{z}) \| p_{t}(\bm{z}))\right)}.
\end{equation}
where $N$ represents the total number of selected source subjects.

The feature-level alignment (FLA) loss between source and target domains is then formulated as:
\begin{align}\label{eq:FLA_ST}
\mathcal{L}^{ST}_{\text{FLA}} &= \sum_{s=1}^N \omega_s D_{\mathrm{CS}}(p_s(\bm{z}) \| p_t(\bm{z}))
\end{align}
where \(\omega_{s}\) are the source weights obtained by Eq.~\eqref{eq:source_weights} and \(N\) is the number of selected sources.

To reduce the heterogeneity among source domains, we also minimize the pairwise CS divergence between source domains:
\begin{equation} \label{eq:FLA_SS}
\begin{aligned}
\mathcal{L}^{SS}_{\text{FLA}} &= \frac{2}{N(N-1)} \sum_{s=1}^N \sum_{\substack{s' > s}}^N \\
&\qquad D_{\mathrm{CS}}(p_s(\bm{z}) \| p_{s'}(\bm{z})),
\end{aligned}
\end{equation}

Harmonizing distributions across source subjects yields a more unified source domain, facilitating better adaptation to the target. The overall FLA loss combines Eq.~\ref{eq:FLA_ST} and Eq.~\ref{eq:FLA_SS}:
\begin{equation} \label{eq:FLA}
\mathcal{L}_{\text{FLA}} = \mathcal{L}^{ST}_{\text{FLA}} + \mathcal{L}^{SS}_{\text{FLA}}.
\end{equation}

\subsubsection{Decision-level alignment}
To ensure optimal discriminative performance, we further align the decision boundaries using the CCS divergence. Similar to the FLA loss, the decision-level alignment (DLA) loss is calculated as:
\begin{equation} \label{eq:DLA_loss}
\begin{aligned}
\mathcal{L}_{\mathrm{DLA}} &= \underbrace{\sum_{s=1}^N \omega_s D_{\mathrm{CCS}}(p_s(y|\bm{z}) \| p_t(y|\bm{z}))}_{\mathcal{L}_{\mathrm{DLA}}^{ST}} \\
&\quad + \underbrace{\frac{2}{N(N-1)} \sum_{s=1}^N \sum_{\substack{s' > s}}^N D_{\mathrm{CCS}}(p_s(y|\bm{z}) \| p_{s'}(y|\bm{z}))}_{\mathcal{L}_{\mathrm{DLA}}^{SS}}.
\end{aligned}
\end{equation}
where \(y\) denotes continuous classification logits produced by the backbone network.

\subsection{Classification and overall loss}
Given one-hot labels \(\mathbf{y}=[y_1, y_2, \dots, y_K]\) for \(K\) MI tasks, the weighted classification loss on source data is computed as:
\begin{equation} \label{eq:classification_loss}
\mathcal{L}_{\mathrm{cls}} = \sum_{s=1}^{N} \omega_s \frac{1}{M_s} \sum_{i=1}^{M_s} \mathrm{CE}\big(g(f(\bm{x}^{s}_i)), y^{s}_i\big),
\end{equation}
where \(\{(\bm{x}^{s}_{i}, y^{s}_{i})\}_{i=1}^{M_s}\) are the labeled samples from source subject \(s\), and \(\mathrm{CE}(\hat{\mathbf{y}}, \mathbf{y}) = -\sum_{c=1}^K y_c \log \hat{y}_c\) denotes the cross-entropy loss between predicted class probabilities \(\hat{\mathbf{y}} = g(f(\bm{x}))\) and one-hot label \(\mathbf{y}\).

The total loss combines all components with dynamic weights \(\alpha_\tau\) and \(\beta_\tau\):
\begin{equation} \label{eq:total_loss}
\mathcal{L}_{\mathrm{total}} = \mathcal{L}_{\mathrm{cls}} + \alpha_\tau \mathcal{L}_{\mathrm{FLA}} + \beta_\tau \mathcal{L}_{\mathrm{DLA}},
\end{equation}
where
\[
\alpha_\tau = \frac{\alpha \exp(-\tau + \tau_0)}{1 + \exp(-\tau + \tau_0)}, \quad \beta_\tau = \frac{\beta}{1 + \exp(-\tau + \tau_0)},
\]
with \(\alpha\) and \(\beta\) as hyperparameters balancing the losses, \(\tau\) the current epoch, and \(\tau_0\) the transition point hyperparameter. The parameter \(\tau_0\) is set to prioritize feature-level alignment in the early training phase and gradually increase the emphasis on decision-level alignment as training progresses. The pseudo-code of the proposed BFM-MSDA framework is given in Algorithm~\ref{alg:bfm-mda}.

\begin{algorithm}[!t]
\footnotesize
\caption{The proposed BFM-MSDA framework}
\label{alg:bfm-mda}
\renewcommand{\algorithmicrequire}{\textbf{Input:}}
\renewcommand{\algorithmicensure}{\textbf{Output:}}
\begin{algorithmic}[1]
\REQUIRE Labeled source EEG \(\{D_s^i\}_{i=1}^S\), unlabeled target \(D_t\), pretrained LaBraM, hyperparameters \(\alpha, \beta\), epochs \(E\)
\ENSURE Predicted labels \(\hat{Y}_t\) for \(D_t\)
\STATE \textbf{Stage 1: Source selection}
\FOR{each source \(i = 1\) to \(S\)}
    \STATE \(\bm{h}_i \gets \text{LaBraM}(D_s^i)\); compute \(d_{s,t}\) by Eq.~\eqref{eq:pairdist}
\ENDFOR
\STATE \(\bm{h}_t \gets \text{LaBraM}(D_t)\); select sources with \(\tilde{w}_{s,t}>1/S\) by Eq.~\eqref{eq:relevanceweight}
\STATE \textbf{Stage 2: MSDA}
\STATE Apply EA (Eq.~\eqref{eq:EA}); initialize \(f\), \(g\)
\FOR{epoch \(\tau = 1\) to \(E\)}
    \STATE Compute \(\omega_s\), \(\mathcal{L}_{\mathrm{FLA}}\), \(\mathcal{L}_{\mathrm{DLA}}\), \(\mathcal{L}_{\mathrm{cls}}\); update by Eq.~\eqref{eq:total_loss}
\ENDFOR
\RETURN \(\hat{Y}_t = g(f(D_t))\)
\end{algorithmic}
\end{algorithm}

\section{Experiments}
\subsection{Datasets and preprocessing}
The effectiveness of the proposed framework is assessed using two benchmark EEG datasets collected under different conditions, with distinct devices, subject groups, and sample sizes.


\subsubsection{Dataset I: BCI Competition IV 2a Dataset}
The BCI Competition IV 2a dataset \cite{brunnerBCICompetition20082008} comprises EEG recordings from nine subjects performing four-class MI tasks involving left hand, right hand, feet, and tongue movements. Data were acquired using 22 Ag/AgCl electrodes positioned according to the standard 10--20 system. Originally sampled at 250 Hz, the signals were subsequently downsampled to 200 Hz and filtered with a 0.5--40 Hz bandpass. For our analysis, we used only left- and right-hand MI trials from each subject's first session, focusing on the 3--6-second post-stimulus window.

\subsubsection{Dataset II: GigaDB Dataset Subset}
\label{subsec:gigadb}
The GigaDB EEG dataset \cite{choEEGDatasetsMotor2017} includes recordings from 52 healthy subjects (s1--s52), each performing right-hand and left-hand MI with 100 trials per class. Each MI trial lasted 3 seconds. EEG signals were recorded from 64 channels at 512 Hz, then downsampled to 200 Hz and band-pass filtered between 0.5 and 40 Hz. Following \cite{zhangMultiSourceGeometricMetric2023}, a subset of 10 subjects ($s2$, $s4$, $s10$, $s13$, $s14$, $s16$, $s17$, $s23$, $s30$, $s31$) is adopted for initial cross-subject leave-one-subject-out (LOSO) evaluation in Sec. \ref{sec:loso_validation}, constituting Dataset II. The full 52-subject GigaDB dataset is further employed for source-selection analysis on a large source base, as described in Sec. \ref{sec:large_pool}.

\subsection{Baseline methods}
The proposed framework is compared against a variety of classic and state-of-the-art methods, including: (1) \textbf{EEG-specific baseline models}: EEGNet \cite{lawhernEEGNetCompactConvolutional2018}, ShallowConvNet \cite{schirrmeisterDeepLearningConvolutional2017}, and EEG Conformer \cite{songEEGConformerConvolutional2023}; (2) \textbf{unsupervised DA techniques}: JAN \cite{longDeepTransferLearning2017}, DJP-MMD \cite{zhangDiscriminativeJointProbability2020}, and DJDAN \cite{hongDynamicJointDomain2021}; (3) \textbf{MSDA methods}: GAT \cite{songGlobalAdaptiveTransformer2023} and ASJDA \cite{liuEnhancingEEGBasedCrossSubject2021}, with EEGNet as the backbone model; and (4) \textbf{foundation-model-based EEG decoders}: LaBraM \cite{jiang_large_2024} and CBraMod \cite{wangCBraModCrissCrossBrain2025}. Following the full-backbone fine-tuning protocol with an MLP classifier \cite{zhang2025multidataset}, the pretrained BFM backbone and a two-layer MLP classification head are jointly fine-tuned on the labeled source subjects under the same LOSO protocol, providing a strong end-to-end BFM decoding baseline.

\subsection{Experimental protocol and setup}
\subsubsection{Cross-subject Evaluation}
\label{sec:loso_validation}
To simulate the cross-subject adaptation scenario, we follow a LOSO evaluation protocol. For Datasets I and II, each subject is sequentially designated as the unlabeled target domain, while the remaining subjects in the dataset serve as potential source domains. Instead of using all available source subjects, the proposed source selection method first identifies those most similar to the target. An adaptive soft-gating rule, as described in Sec.~\ref{subsec:source_selection}, is used to identify relevant source subjects for each target. The adaptive soft-gated selection is the default operating mode of the proposed framework.

To test the BFM-guided source selection independently, we additionally train the backbone on each ordered source--target pair without adaptation and correlate BFM distance with that single-source accuracy, reported in Sec.~\ref{sec:correlation} and~\ref{app:correlation}.

\subsubsection{Controlled source-selection analysis under a large source base}
\label{sec:large_pool}
To further validate the necessity and practicality of source selection in a large source pool, we conducted an additional controlled analysis using the full 52-subject GigaDB dataset, simulating a real-world scenario in which a large database of source subjects is available. In this experiment, the 10 subjects from Dataset II are treated as target domains, while the remaining 42 subjects from the GigaDB dataset serve as potential source domains. For this mechanism study, we apply a fixed 25th-percentile threshold to the BFM-derived source-target distances, resulting in approximately 10 to 13 selected source subjects per target. DA is then conducted exclusively using these selected sources. For comparison, we conduct parallel experiments with 5 randomly selected groups of 12 source subjects to quantify the benefit of informed source filtering under a matched source-set size. Additionally, we vary the fixed percentile threshold from the 5th to the 50th percentile to examine the trade-off among the number of selected sources, classification accuracy, and computational cost.

EEGNet is used as the backbone network for DA experiments, with its temporal kernel size set at 100, corresponding to half the EEG sampling rate \cite{lawhernEEGNetCompactConvolutional2018}. The balancing factors for the losses were fixed at \(\alpha = 0.7\) and \(\beta = 1.4\) for both datasets. The transition point \(\tau_0 = 100\) is set to balance early-stage feature alignment with later-stage decision alignment. For optimization, we employ the Adam optimizer with a learning rate of 0.001 and a batch size of 32 for both datasets. To enhance training stability and convergence, we incorporate the Cosine Annealing learning rate scheduler. The number of training epochs is set at 500 for Dataset I and 300 for Dataset II. The bandwidth \(\sigma\) of the Gaussian kernels used to estimate CS divergence is determined by the median heuristic on pairwise distances within each mini-batch. All experiments are implemented in Python with PyTorch and executed on an NVIDIA RTX 4070 Ti GPU.

The classification accuracy and Cohen's kappa are used to evaluate the models' classification performance. Kappa is defined as follows:
\begin{equation}
\kappa = \frac{p_o - p_e}{1 - p_e}
\end{equation}
where \(p_o\) represents the observed accuracy and \(p_e\) denotes the accuracy by chance. All methods are trained and evaluated in 10 repeated runs with random initialization.

\section{Results}
\subsection{Cross-subject classification performances}
The cross-subject classification accuracies and kappa values on Dataset I and Dataset II are summarized in Tables~\ref{tab:dataset1} and~\ref{tab:dataset2}, respectively. Wilcoxon signed-rank tests were performed to compare the proposed method with each baseline. Overall, the proposed BFM-MSDA significantly outperformed most baselines ($p < 0.05$) in average performance on both datasets, demonstrating superior generalization across subjects. On Datasets I and II, our method achieves the best performance on most target subjects. The improvement is particularly notable for the subjects who originally exhibit lower decoding performance on the baseline model.

Compared to classical MI decoding models such as EEGNet, ShallowConvNet, and Conformer, which do not explicitly address domain shifts, our method's DA strategy specifically addresses inter-subject variability. MMD-based methods, such as JAN and DJP-MMD, improve transferability by aligning marginal distributions but lack explicit alignment of conditional distributions, thereby limiting their performance gains. Adversarial learning-based approaches such as DJDAN and GAT implicitly align domain features and improve class discriminability, but can be unstable and sensitive to hyperparameters. The results highlight the advantage of selectively leveraging relevant source subjects and conducting both feature-level and decision-level alignment to mitigate inter-subject variability.

As for the BFM-based decoders, full-backbone fine-tuning of LaBraM and CBraMod yields competitive cross-subject accuracies (82.57\% and 84.10\% on Dataset~I; 72.63\% and 74.28\% on Dataset~II), confirming the benefit of task-specific adaptation of large pretrained EEG models. Nevertheless, the proposed BFM-MSDA still significantly outperforms both fine-tuned BFMs ($p < 0.01$), while relying on a lightweight EEGNet backbone and using the BFM only as a frozen embedding engine for source selection. This suggests that target-aware multi-source alignment can complement---and in this setting surpass---direct fine-tuning of foundation-model decoders.

\begin{table*}[!t]
\centering
\caption{Cross-subject accuracies (mean \(\pm\) std.\ in \%) and kappa on Dataset I. Best performances are highlighted in bold. \(^{*}\) represents \(p<0.05\) and \(^{**}\) represents \(p<0.01\).}
\label{tab:dataset1}
\begingroup\small\setlength{\tabcolsep}{4pt}\renewcommand{\arraystretch}{1.0}%
\begin{tabular}{@{}lcccccccccrr@{}}
\toprule
\multirow{2}{*}{Methods} & \multicolumn{9}{c}{Subject} & \multirow{2}{*}{Avg. acc.} & \multirow{2}{*}{Kappa} \\
\cmidrule(lr){2-10}
& 1 & 2 & 3 & 4 & 5 & 6 & 7 & 8 & 9 &  &  \\
\midrule
EEGNet & 72.92 & 64.58 & 79.17 & 78.47 & 86.11 & 74.31 & 83.33 & 79.86 & 82.64 & 77.93$^{**}$ $\pm$ 6.52 & 0.5586 \\
ShallowConvNet & 75.73 & 74.89 & 78.47 & 76.39 & 81.25 & 85.43 & 87.50 & 83.33 & 76.39 & 79.93$^{**}$ $\pm$ 4.62 & 0.5986 \\
Conformer & 81.59 & 68.53 & 88.07 & 82.04 & 74.68 & 78.54 & 86.97 & 82.59 & 83.74 & 80.75$^{**}$ $\pm$ 6.10 & 0.6150 \\
JAN & 79.69 & 76.32 & 84.25 & 75.69 & 86.81 & 84.72 & 89.19 & 90.28 & 81.94 & 83.21$^{*}$ $\pm$ 5.25 & 0.6642 \\
DJP-MMD & 79.78 & 72.22 & 89.58 & 81.94 & 85.42 & 83.33 & 88.89 & 89.58 & 86.28 & 84.11$^{*}$ $\pm$ 5.64 & 0.6823 \\
GAT & 82.42 & 75.69 & \textbf{92.06} & 82.78 & 88.28 & 81.34 & 82.06 & 85.42 & 81.94 & 83.55$^{**}$ $\pm$ 4.63 & 0.6711 \\
ASJDA & 79.22 & 74.61 & 89.58 & 78.08 & 82.64 & 82.81 & 85.42 & \textbf{90.97} & 86.81 & 83.35$^{*}$ $\pm$ 5.42 & 0.6670 \\
DJDAN & 81.25 & 76.39 & 91.36 & 78.47 & 81.25 & \textbf{86.19} & 92.36 & 82.64 & 84.72 & 83.85$^{*}$ $\pm$ 5.42 & 0.6770 \\
LaBraM & 80.56 & \textbf{78.17} & 85.19 & 80.81 & 79.76 & 80.26 & 89.81 & 85.71 & 82.87 & 82.57$^{**}$ $\pm$ 3.69 & 0.6514 \\
CBraMod & 79.32 & 74.38 & 88.65 & 80.92 & 86.81 & 82.43 & 90.84 & 88.35 & 85.24 & 84.10$^{**}$ $\pm$ 5.28 & 0.6821 \\
\textbf{Proposed} & \textbf{83.19} & 76.53 & 91.74 & \textbf{83.26} & \textbf{89.37} & 81.46 & \textbf{92.50} & 90.14 & \textbf{87.29} & \textbf{86.16} $\pm$ 5.37 & \textbf{0.7233} \\
\bottomrule
\end{tabular}\endgroup
\end{table*}
\begin{table*}[!t]
\centering
\caption{Cross-subject accuracies (mean \(\pm\) std.\ in \%) and kappa on Dataset II. Best performances are highlighted in bold. \(^{*}\) represents \(p<0.05\) and \(^{**}\) represents \(p<0.01\).}
\label{tab:dataset2}
\begingroup\small\setlength{\tabcolsep}{4pt}\renewcommand{\arraystretch}{1.0}%
\begin{tabular}{@{}lcccccccccccc@{}}
\toprule
\multirow{2}{*}{Methods} & \multicolumn{10}{c}{Subject} & \multirow{2}{*}{Avg. acc.} & \multirow{2}{*}{Kappa} \\
\cmidrule(lr){2-11}
& 1 & 2 & 3 & 4 & 5 & 6 & 7 & 8 & 9 & 10 &  &  \\
\midrule
EEGNet & 66.57 & 60.96 & 82.70 & 70.14 & 73.89 & 59.05 & 55.68 & 63.56 & 88.29 & 65.06 & 68.59$^{**}$ $\pm$ 10.41 & 0.3718 \\
ShallowConvNet & 73.92 & 67.63 & 75.05 & 71.55 & 75.23 & 63.65 & 70.56 & 67.70 & 78.02 & 66.37 & 70.97$^{**}$ $\pm$ 4.60 & 0.4194 \\
Conformer & 77.62 & 71.01 & 78.80 & 74.12 & 78.99 & 66.83 & 74.09 & 71.09 & 81.92 & 69.69 & 74.42$^{**}$ $\pm$ 4.82 & 0.4883 \\
JAN & 78.27 & 71.38 & 80.72 & 73.75 & 79.46 & 76.84 & 73.05 & 70.30 & 87.35 & 70.28 & 76.14$^{**}$ $\pm$ 5.47 & 0.5228 \\
DJP-MMD & 76.31 & \textbf{74.43} & 78.26 & 71.85 & 73.62 & 77.94 & 70.22 & 71.43 & 82.45 & 71.52 & 74.80$^{**}$ $\pm$ 3.89 & 0.4961 \\
GAT & 76.25 & 71.36 & 83.73 & 69.32 & 77.45 & 75.73 & \textbf{74.84} & 72.84 & 90.45 & 73.64 & 76.56$^{*}$ $\pm$ 6.24 & 0.5312 \\
ASJDA & 79.03 & 69.72 & 78.16 & 69.52 & 67.46 & 72.56 & 66.68 & \textbf{76.50} & 89.15 & 72.42 & 74.12$^{**}$ $\pm$ 6.81 & 0.4824 \\
DJDAN & 78.36 & 70.87 & 80.53 & 72.73 & 74.67 & \textbf{79.27} & 71.38 & 72.54 & 90.12 & 70.94 & 76.14$^{*}$ $\pm$ 6.09 & 0.5228 \\
LaBraM & 75.06 & 66.82 & 77.21 & 70.36 & 74.84 & 72.34 & 68.89 & 72.55 & 79.61 & 68.58 & 72.63$^{**}$ $\pm$ 4.08 & 0.4525 \\
CBraMod & 74.37 & 70.09 & 79.98 & 68.45 & 74.96 & 73.58 & 72.87 & 71.27 & 85.36 & 71.91 & 74.28$^{**}$ $\pm$ 4.99 & 0.4857 \\
\textbf{Proposed} & \textbf{80.52} & 73.53 & \textbf{85.15} & \textbf{74.48} & \textbf{81.42} & 75.51 & 72.05 & 73.15 & \textbf{92.75} & \textbf{75.52} & \textbf{78.41} $\pm$ 6.57 & \textbf{0.5682} \\
\bottomrule
\end{tabular}\endgroup
\end{table*}

\subsection{Controlled analysis of source selection under large source base}
To analyze the effect of source selection in scenarios with a large pool of candidate source subjects, we conducted a controlled study on the full GigaDB dataset. Each of the 10 subjects from Dataset II was treated as a target domain, while the remaining 42 subjects served as potential source domains. Table~\ref{tab:source_groups} compares classification accuracies obtained using this informed source filtering against those from randomly selected source subsets of similar size. The results show that informed source filtering consistently yields stronger performance across most target subjects.

\begin{figure}[!t]
\centering
\includegraphics[width=\columnwidth]{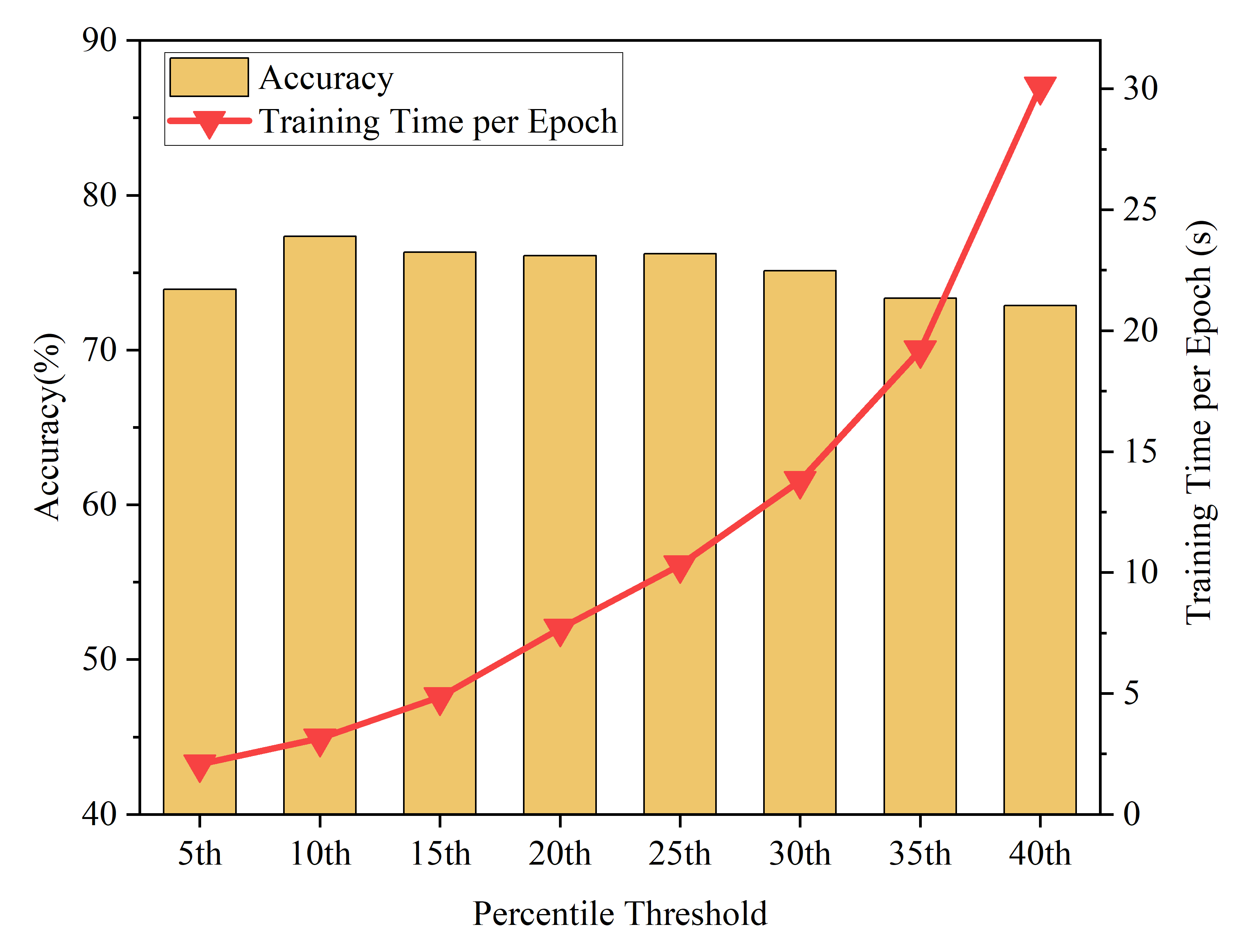}
\caption{Effect of selection threshold on performance and training time.}
\label{fig:computationalcost}
\end{figure}

As illustrated in Fig.~\ref{fig:computationalcost}, we further varied the fixed percentile threshold from the 5th to the 50th percentile to analyze the trade-off among the number of selected sources, classification accuracy, and training time per epoch. Accuracy peaks at 77.34\% under the 25th-percentile threshold used in Table~\ref{tab:source_groups}. A tighter 5th-percentile cutoff discards useful sources and lowers accuracy; a looser 50th-percentile cutoff drops accuracy to 72.56\% and raises training time from 3.14\,s to 66.32\,s per epoch. This empirical training cost is consistent with the analysis in~\ref{app:complexity}.

These findings highlight the importance of proper source selection in MSDA for EEG decoding, especially when scaling to large datasets with numerous potential sources. By effectively filtering out irrelevant subjects, the proposed framework enables faster model tuning and more practical deployment in real-world BCI applications.

\begin{table*}[!t]
\centering
\caption{Cross-subject performances of different source groups from the GigaDB dataset. Best performances are highlighted in bold. \(^{*}\) represents \(p<0.05\) and \(^{**}\) represents \(p<0.01\).}
\label{tab:source_groups}
\begingroup\small\setlength{\tabcolsep}{4pt}\renewcommand{\arraystretch}{1.0}%
\begin{tabular}{@{}lccccccccccr@{}}
\toprule
\multirow{2}{*}{Group} & \multicolumn{10}{c}{Subject} & \multirow{2}{*}{Avg. acc.} \\
\cmidrule(lr){2-11}
& 1 & 2 & 3 & 4 & 5 & 6 & 7 & 8 & 9 & 10 &  \\
\midrule
Random Group 1 & 66.42 & 63.57 & 80.65 & 66.50 & 79.51 & 67.83 & 59.52 & 66.36 & 89.52 & 56.23 & 69.61$^{**}$ $\pm$ 10.37 \\
Random Group 2 & 73.85 & \textbf{74.31} & 79.27 & 73.03 & 75.65 & 70.95 & 71.18 & 73.54 & 85.42 & 74.85 & 75.21$^{*}$ $\pm$ 4.29 \\
Random Group 3 & 72.38 & 70.15 & 71.33 & 72.27 & 72.92 & 73.65 & \textbf{78.63} & 71.81 & 88.71 & 74.75 & 74.66$^{*}$ $\pm$ 5.45 \\
Random Group 4 & 71.38 & 72.81 & 79.52 & 65.15 & 78.81 & 71.57 & 69.82 & 70.55 & 89.17 & 65.67 & 73.44$^{**}$ $\pm$ 7.24 \\
Random Group 5 & 72.94 & 67.35 & 73.17 & 67.24 & 75.64 & 68.32 & 70.72 & \textbf{74.36} & 88.46 & 66.75 & 72.50$^{**}$ $\pm$ 6.47 \\
\textbf{Proposed} & \textbf{77.36} & 70.87 & \textbf{81.56} & \textbf{74.24} & \textbf{80.73} & \textbf{74.02} & 74.31 & 72.35 & \textbf{91.70} & \textbf{76.26} & \textbf{77.34} $\pm$ 6.09 \\
\bottomrule
\end{tabular}\endgroup
\end{table*}

\subsection{BFM distance and single-source transferability}
\label{sec:correlation}
Stage~1 of our proposed framework selects sources by BFM-space distance. To directly test the premise underlying this design, we evaluate each ordered source--target pair using a backbone trained on source \(s\) alone and evaluated on target \(t\), without DA. This single-source transferability isolates the intrinsic utility of a candidate source from interactions among sources and from the downstream alignment objective \cite{murtazaNoveltyNeuralSignatures2025}. Full experimental and statistical-testing details are provided in~\ref{app:correlation}.

Formally, the transferability of an ordered pair is
\begin{equation}
A(s \rightarrow t) = \operatorname{acc}\bigl(f_{\theta(s)},\; \mathcal{D}_t\bigr),
\qquad s \neq t,
\label{eq:transferability}
\end{equation}
where \(f_{\theta(s)}\) denotes the backbone trained exclusively on labelled data from source \(s\).

Because target subjects vary substantially in their overall transfer difficulty, pooling all pairs would confound source relevance with target-specific performance. We therefore assess the association within each target. Let \(\rho_t\) be Spearman's rank correlation coefficient between BFM distance and single-source accuracy across the candidate sources of target \(t\). The summary statistic is the mean of these coefficients,
\[
\bar{\rho} = \frac{1}{T}\sum_{t=1}^{T}\rho_t,
\]
where \(T\) is the number of targets (Table~\ref{tab:corr_summary}). The mean correlation is negative and significant on both datasets: \(\bar{\rho}=-0.304\) (\(p=.016\)) on Dataset~I and \(\bar{\rho}=-0.265\) (\(p=.018\)) on Dataset~II. Thus, after controlling for target difficulty, closer BFM embeddings reliably identify source subjects with higher transferability. This result provides direct empirical support for using BFM embedding geometry as the retrieval criterion before multi-source adaptation.

\begin{table}[!t]
\centering
\caption{Mean within-target Spearman correlation \(\bar{\rho}\) between BFM distance and single-source transferability. \(\bar{\rho}\) is the average of the per-target rank correlations \(\rho_t\). The \(p\)-value is from a blocked permutation test of \(\bar{\rho}\).}
\label{tab:corr_summary}
\begingroup\renewcommand{\arraystretch}{0.9}%
\begin{tabular}{@{}lrrr@{}}
\toprule
Dataset & Mean Spearman \(\bar{\rho}\) & \(p\)-value & Targets \\
\midrule
Dataset I & \(-0.304\) & \(0.016\) & 9 \\
Dataset II & \(-0.265\) & \(0.018\) & 10 \\
\bottomrule
\end{tabular}\endgroup
\end{table}

\begin{figure*}[!t]
  \centering
  \subfloat[]{\includegraphics[width=0.48\textwidth]{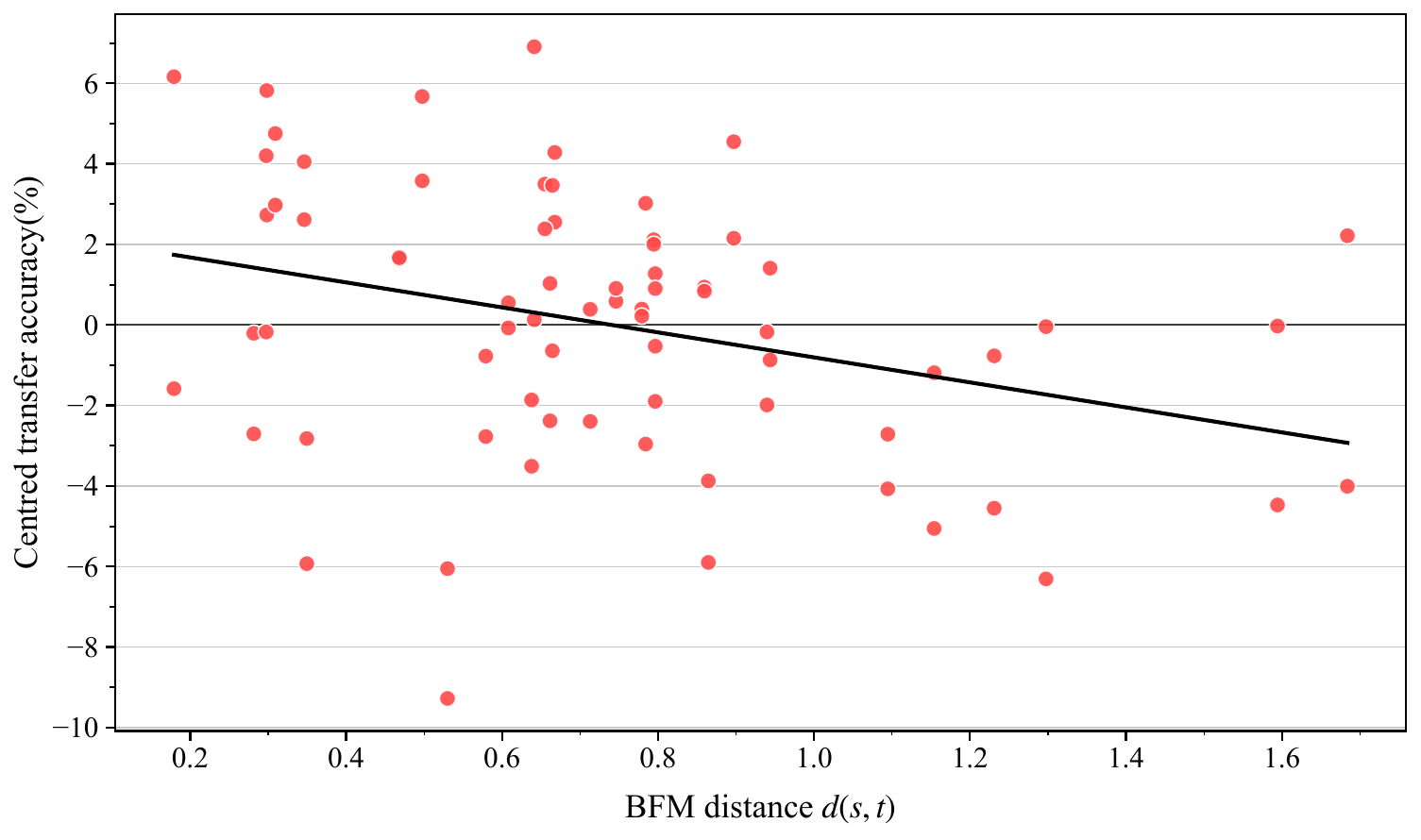}\label{fig:corr_dataset1}}
  \hfill
  \subfloat[]{\includegraphics[width=0.48\textwidth]{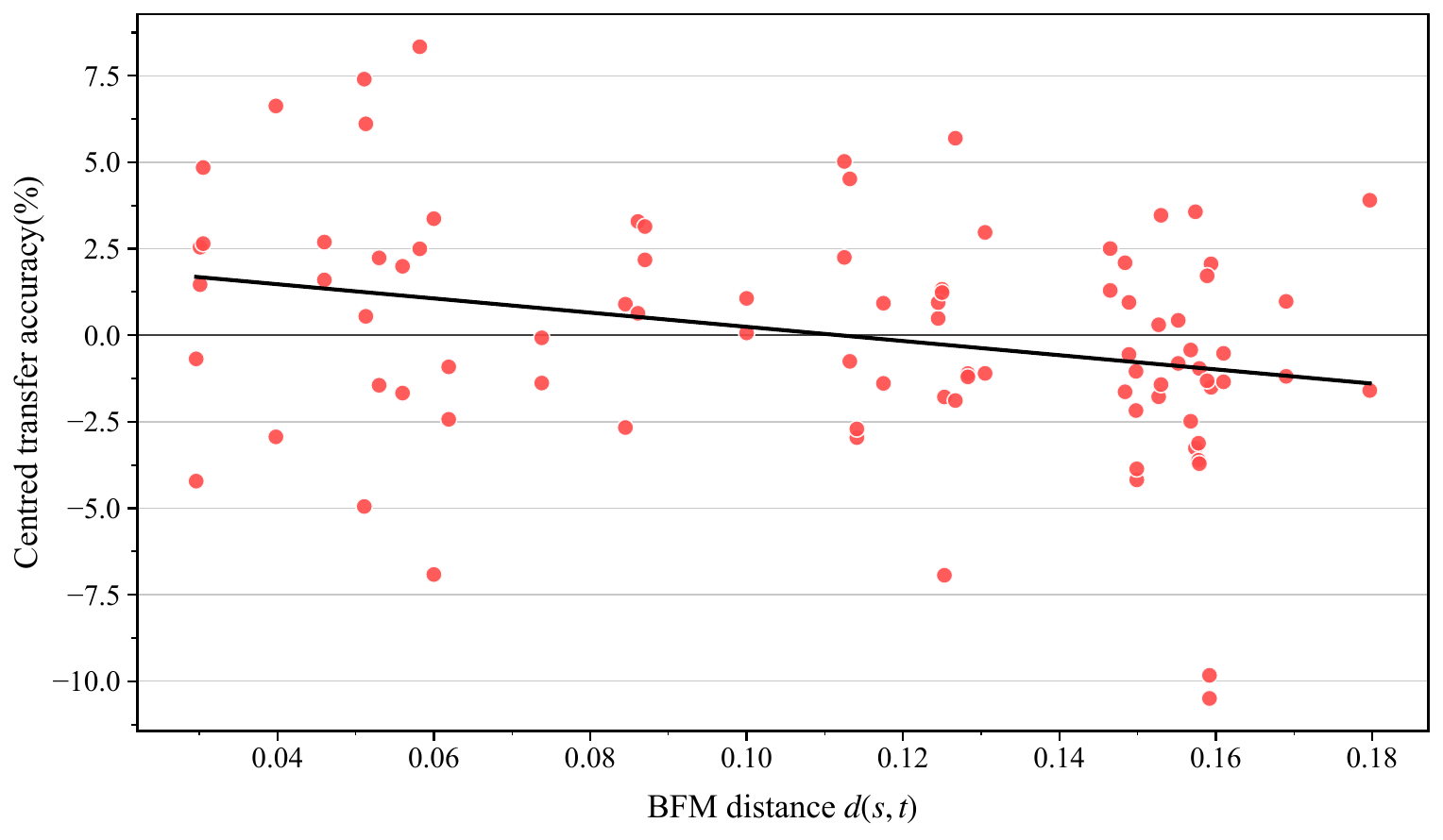}\label{fig:corr_dataset2}}
  \caption{BFM distance against single-source transfer accuracy after centring accuracy within each target. Each point represents an ordered source--target pair, and the solid line is a least-squares fit. Larger BFM distance is associated with lower transferability on both (a) Dataset~I and (b) Dataset~II.}
  \label{fig:corr}
\end{figure*}

\subsection{Ablation experiments}
\label{sec:ablation}
To evaluate the effectiveness of the key components in the proposed framework, we conducted ablation experiments on both datasets under consistent experimental settings. The main contributions of our framework are the BFM-empowered source selection module and the MSDA leveraging CS divergences. Accordingly, we compared the following five variants:
\begin{enumerate}
\item Baseline EEGNet without DA or source selection.
\item DA with DLA only.
\item DA with FLA only.
\item DA incorporating both FLA and DLA, but without source selection.
\item Full BFM-MSDA framework, including source selection and MSDA.
\end{enumerate}

\begin{figure*}[!t]
\centering
\subfloat[]{\includegraphics[width=0.48\textwidth]{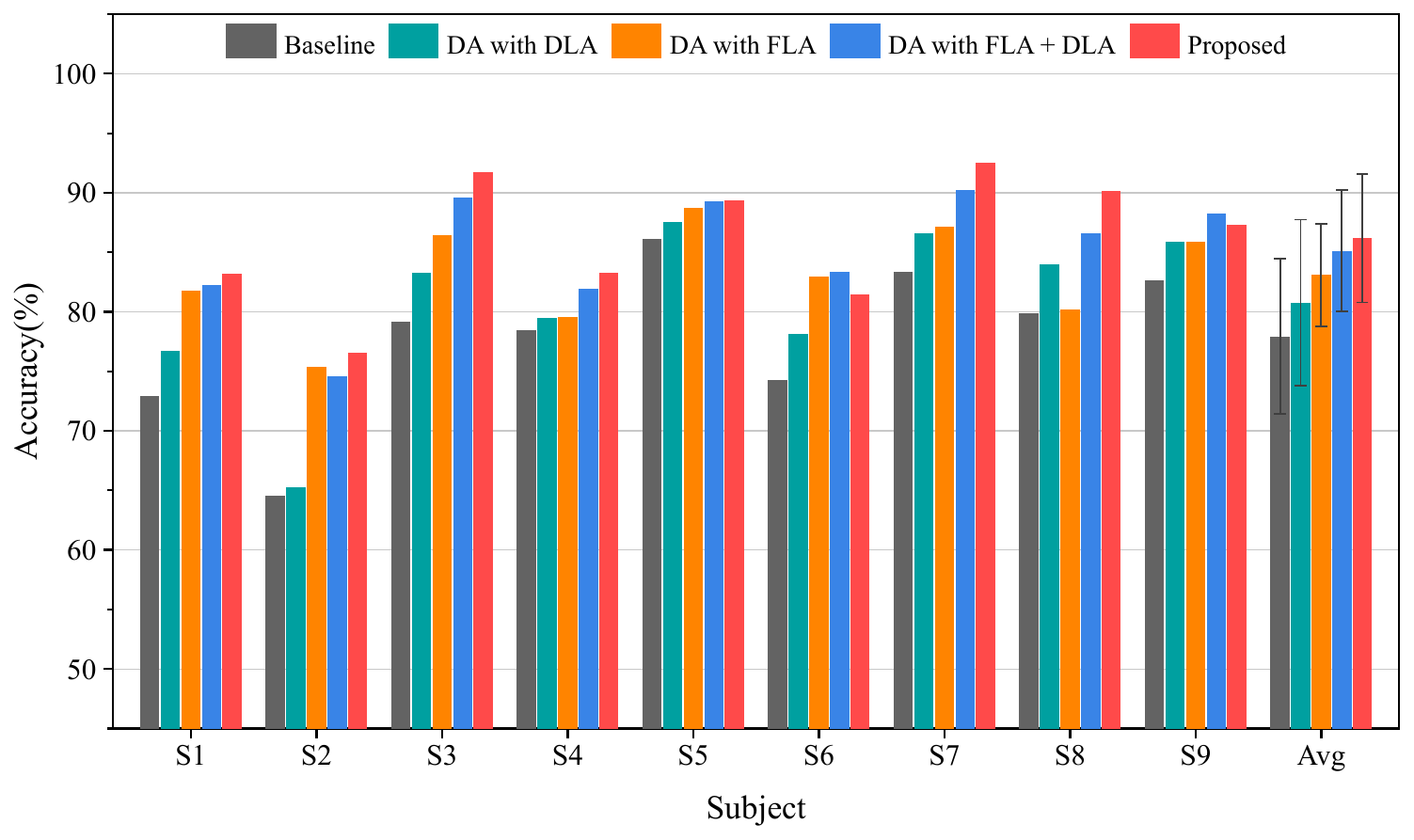}\label{ablation_2a}}
\hfill
\subfloat[]{\includegraphics[width=0.48\textwidth]{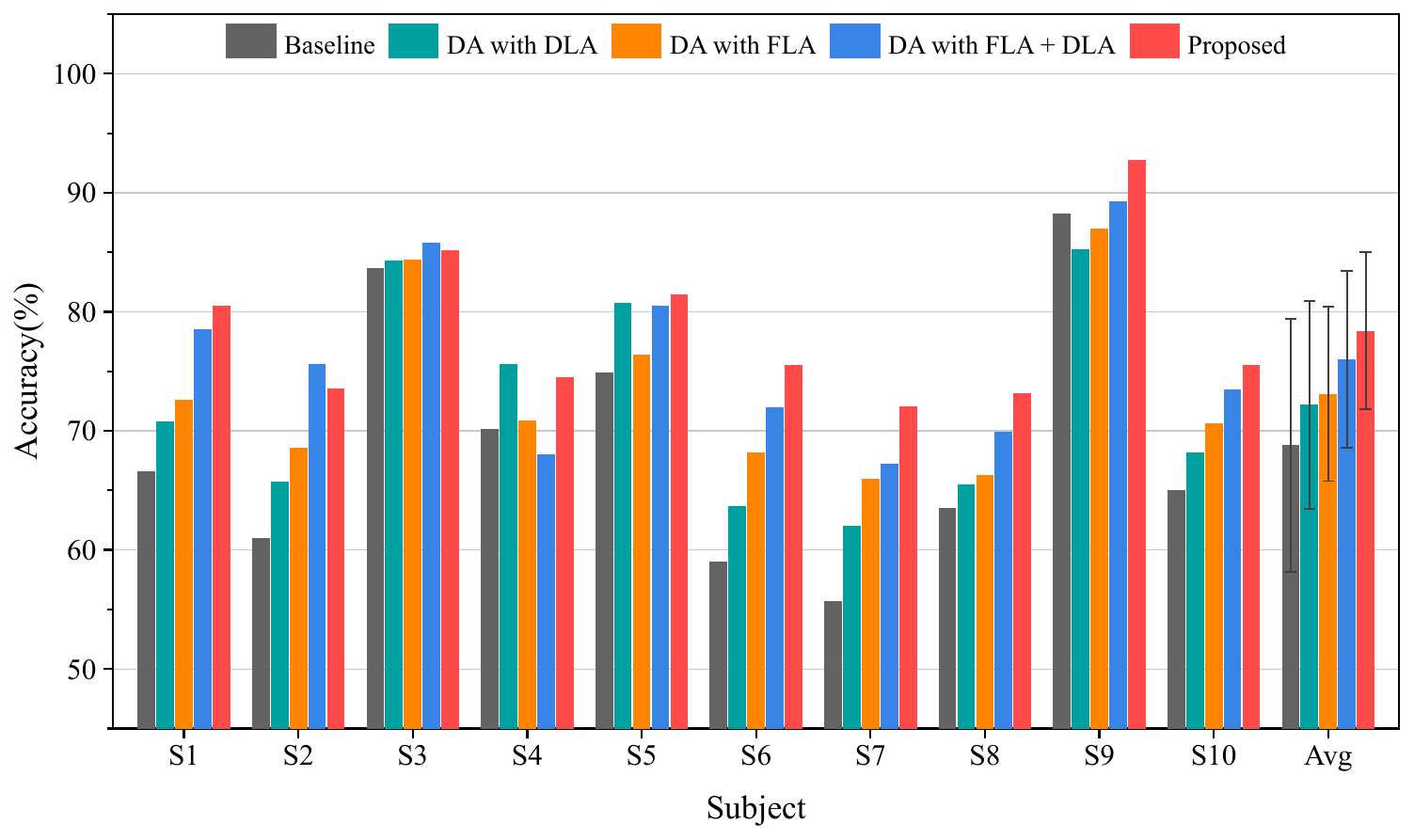}\label{ablation_cho}}
\caption{Cross-subject classification accuracies of ablation variants. (a) Dataset I. (b) Dataset II.}
\label{fig_ablation}
\end{figure*}

The cross-subject classification accuracies of each variant on both datasets are presented in Fig.~\ref{fig_ablation}. Both alignment terms improve over the baseline when applied alone: DLA raises the average accuracy by 2.83\% on Dataset I and 3.79\% on Dataset II, while FLA yields larger gains of 5.17\% and 4.31\%, respectively. Combining the two terms reaches 85.12\% on Dataset I and 76.04\% on Dataset II, for cumulative gains of 7.19\% and 7.45\%. These joint gains fall short of the sum of the two individual contributions, indicating complementary rather than fully independent effects. Source selection then adds a further 1.04\% on Dataset I and 2.37\% on Dataset II. The DLA and source-selection increments are larger on Dataset II, which exhibits greater inter-subject variability. Overall, the full framework improves over the baseline by 8.23\% on Dataset I and 9.82\% on Dataset II. The gains are larger on the more difficult targets: averaged over the three lowest-baseline subjects of each dataset, the improvement is 9.79\% on Dataset I and 15.13\% on Dataset II.

When the same BFM ranking is evaluated without domain adaptation, it remains above random selection at every source-set size below the full pool, accuracy rises and then falls as more distant sources are admitted, and the cost of using the entire pool is higher than the selection increment inside the complete model (~\ref{app:k_curves}).

\subsection{Comparative experiments with KL divergence}
\label{sec:kl_comparison}
KL divergence can also be estimated non-parametrically. Therefore, we conducted comparative experiments replacing the CS/CCS divergence losses with the joint KL divergence defined by Eq.~\ref{eq:joint_kl_forward} in ~\ref{app:kde_kl}. All training schedules, optimizers, and hyperparameters were kept identical to the CS-based configuration to ensure a fair comparison.

As demonstrated in Table~\ref{tab:KL_comparison}, the KDE-based KL variant achieves higher accuracy than the baseline but fails to match the performance of the proposed CS/CCS method on both datasets. In practice, we observed that the KL estimator was numerically less stable with bandwidth selection.

\begin{table}[!t]
\centering
\caption{Cross-subject performance comparison between CS and KDE-based KL divergences.}
\label{tab:KL_comparison}
\begingroup\renewcommand{\arraystretch}{0.9}%
\resizebox{\columnwidth}{!}{%
\begin{tabular}{@{}lccc@{}}
\toprule
Dataset & Baseline & CS/CCS & KDE-KL \\
\midrule
Dataset I & 77.93 $\pm$ 6.52 & \textbf{86.16} $\pm$ 5.37 & 79.34 $\pm$ 8.46 \\
Dataset II & 68.59 $\pm$ 10.41 & \textbf{78.41} $\pm$ 6.57 & 71.37 $\pm$ 9.24 \\
\bottomrule
\end{tabular}}\endgroup
\end{table}

CS divergence provides a provably tighter generalization bound than KL divergence \cite{yin_domain_2024}, and enjoys superior numerical stability by avoiding logarithmic singularities. These properties explain the consistent advantages of our CS/CCS formulation over the KDE-based KL estimator in DA.

\subsection{Visualization}

\subsubsection{Inter-subject similarity via MDS and heat maps}
\begin{figure*}[!t]
\centering
\subfloat[]{\includegraphics[width=0.45\textwidth]{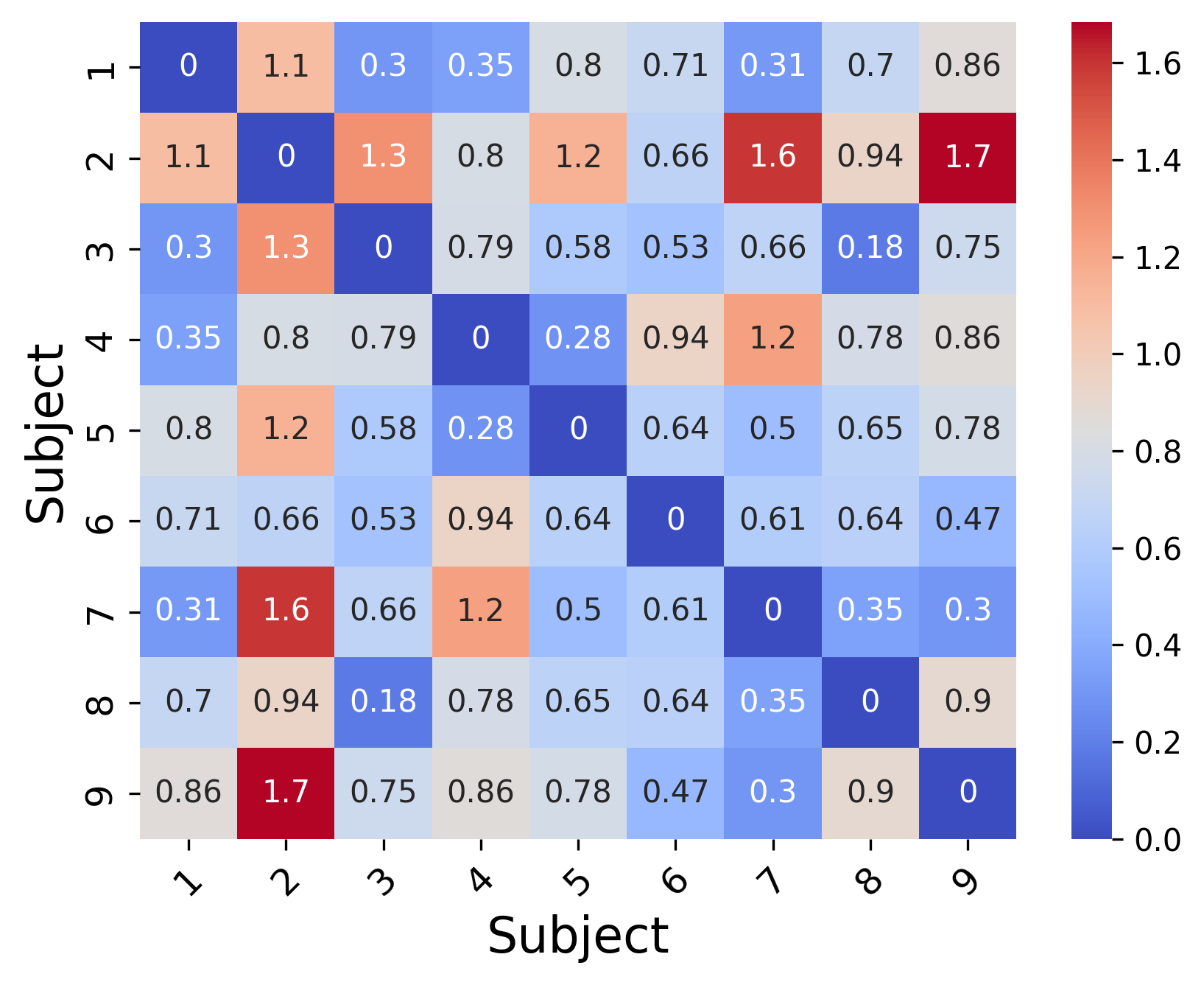}\label{heatmap_2a}}
\hfill
\subfloat[]{\includegraphics[width=0.45\textwidth]{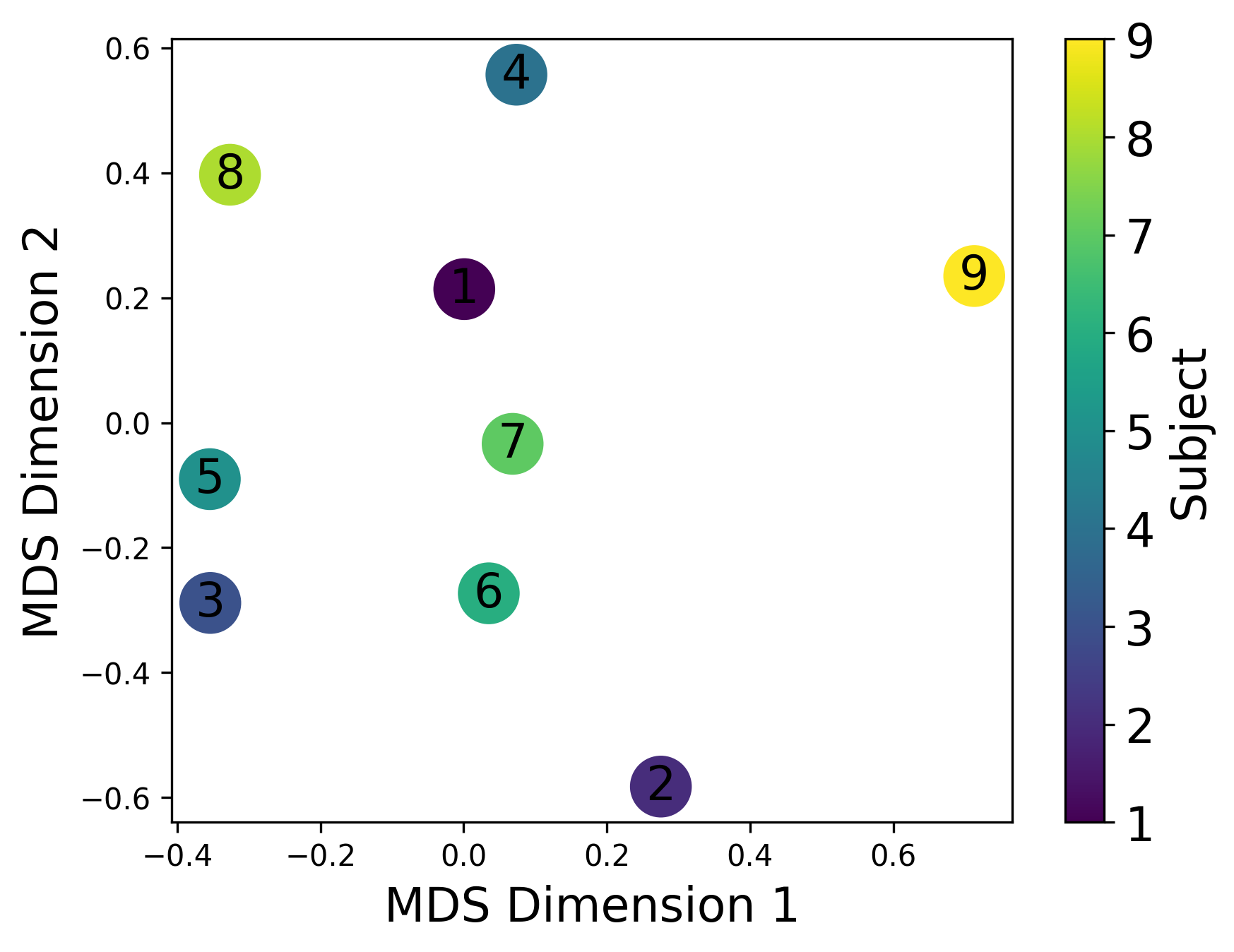}\label{MDS_2a}}\\
\subfloat[]{\includegraphics[width=0.45\textwidth]{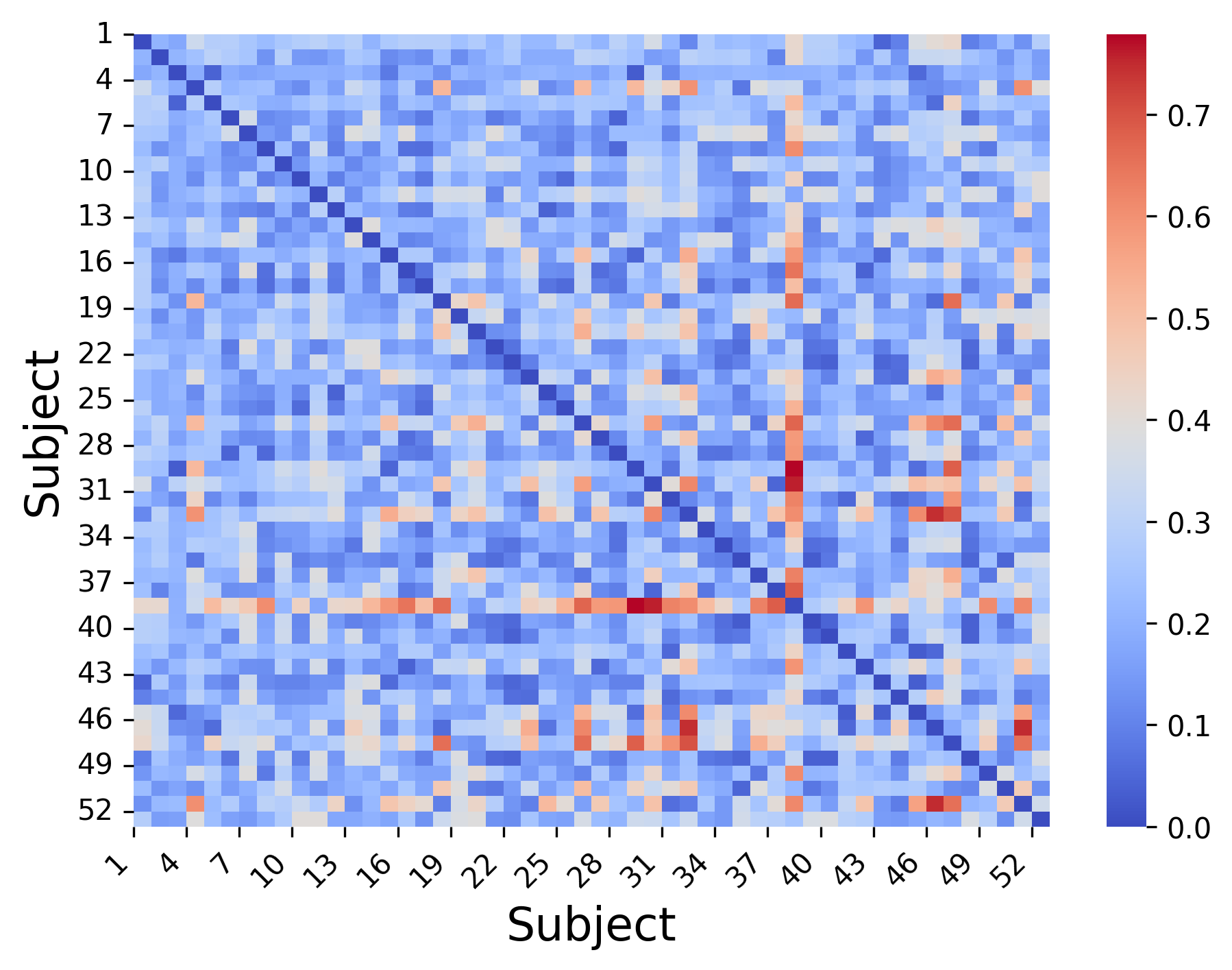}\label{heatmap_cho}}
\hfill
\subfloat[]{\includegraphics[width=0.45\textwidth]{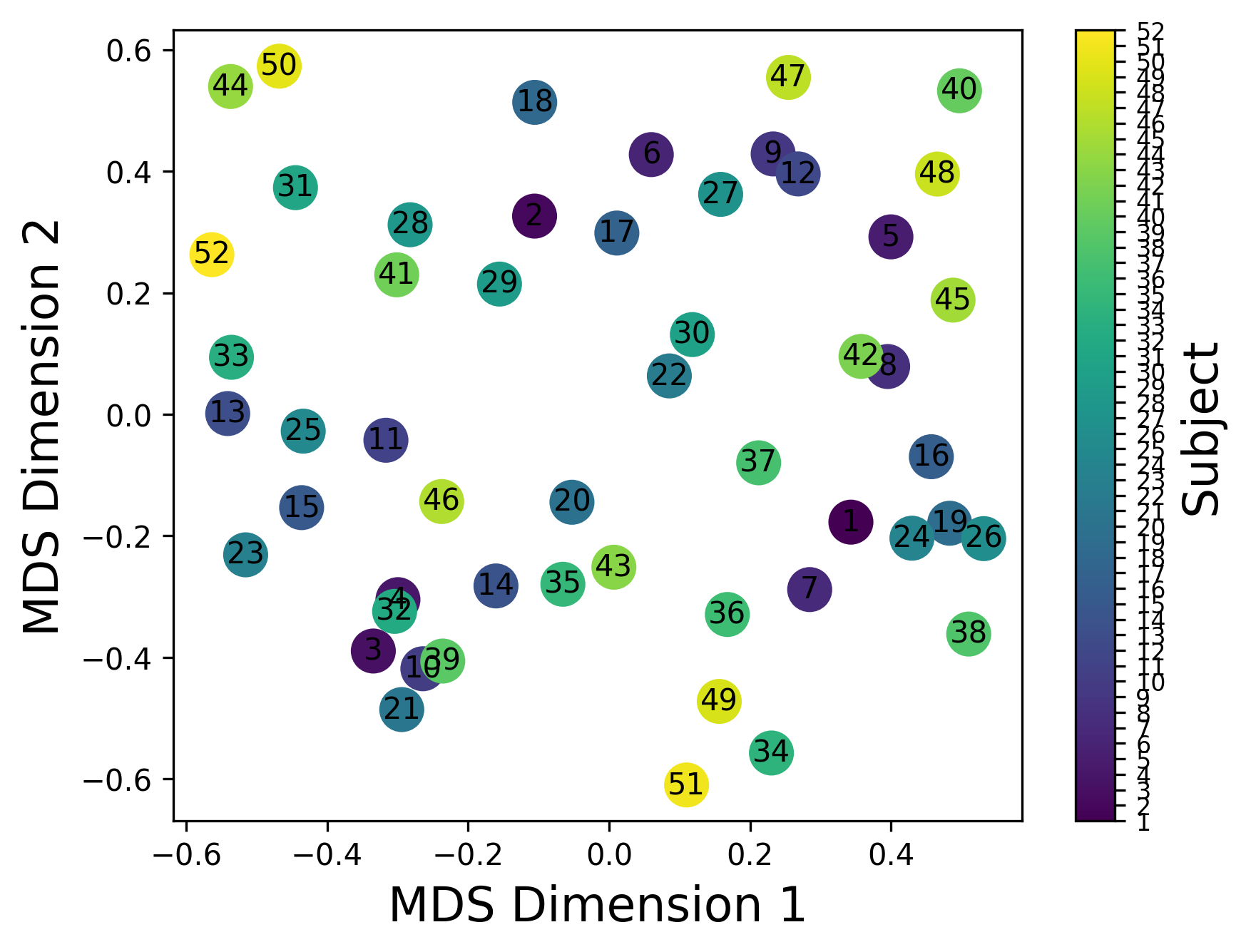}\label{MDS_cho}}
\caption{Heat maps and MDS visualizations of pairwise distances for Datasets I and II. (a) Heat map of inter-subject distances for Dataset I. (b) MDS visualization of inter-subject distances for Dataset I. (c) Heat map of inter-subject distances for Dataset II. (d) MDS visualization of inter-subject distances for Dataset II.}
\label{fig_distance}
\end{figure*}

Multidimensional Scaling (MDS) \cite{weinbergerUnsupervisedLearningImage2006} has been employed to visualize the proposed source selection process. MDS projects the high-dimensional distance matrix into a two-dimensional space, facilitating visual inspection of the relative similarities among subjects. Figs.~\ref{heatmap_2a} and \ref{MDS_2a} show the heat map and MDS visualization of inter-subject distances for Dataset I, respectively, while Figs.~\ref{heatmap_cho} and \ref{MDS_cho} present the corresponding heat map and MDS plot for Dataset II. The MDS plots reveal clusters of subjects with similar EEG feature distributions. Together with the within-target analysis in Sec.~\ref{sec:correlation}, this geometry is associated with higher single-source transfer.

\subsubsection{Feature discriminability on target domain via t-SNE}
\begin{figure*}[!t]
\centering
\includegraphics[width=0.85\textwidth]{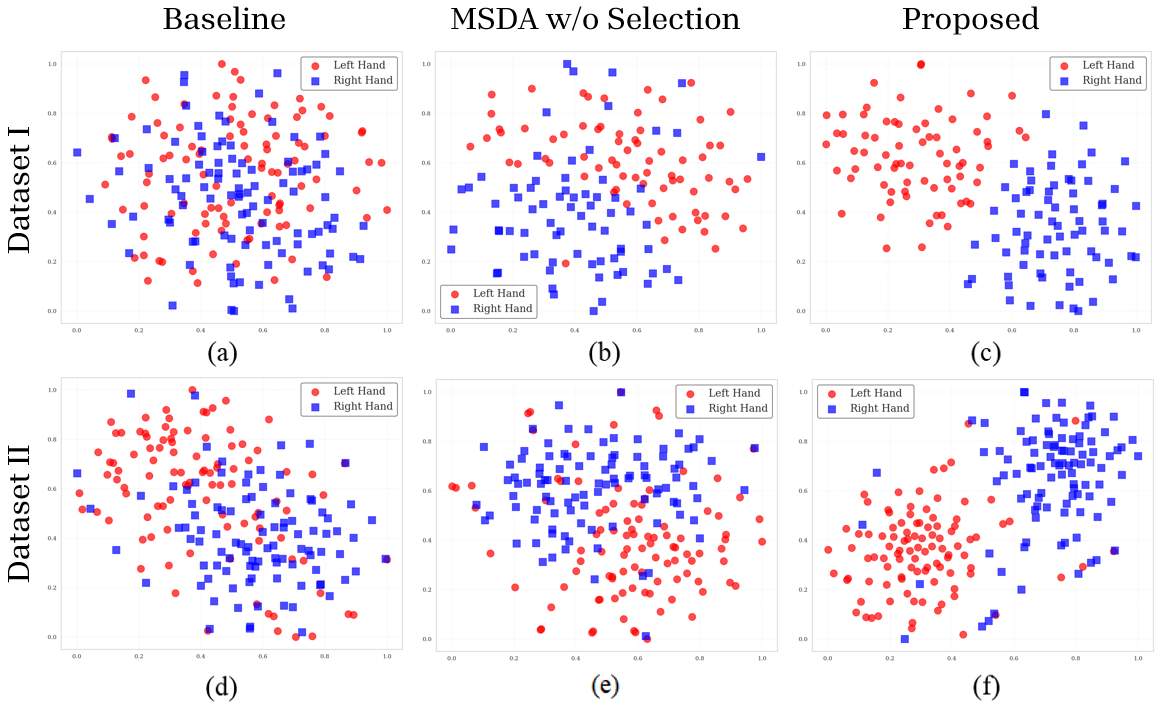}
\caption{t-SNE visualization of target domain features extracted by the feature extractor for subject 1 from Dataset I (top row) and Dataset II (bottom row). (a) and (d) baseline EEGNet without DA, (b) and (e) MSDA without source selection, and (c) and (f) full BFM-MSDA framework.}
\label{fig:t-SNE}
\end{figure*}

We visualized the latent features extracted by the feature extractor using t-distributed stochastic neighbor embedding (t-SNE), taking subject 1 from Datasets I and II as examples. Figs.~\ref{fig:t-SNE} (a)--(f) present the t-SNE embeddings of target domain features for three cases on both datasets: the baseline EEGNet model without DA, the EEGNet enhanced with MSDA but without source selection, and the full BFM-MSDA framework.

From the visualizations, it is evident that the baseline model produces feature embeddings with significant overlap between different MI classes, indicating poor class separability and limited generalization to the target domain. Incorporating MSDA without source selection improves feature clustering and separation, yielding more distinct clustering corresponding to different MI classes. The introduction of source selection further refines this separation, yielding clearly defined and well-separated clusters for each MI class. The tighter, more compact clusters in (c) and (f) indicate that the learned feature representations are both more discriminative and better adapted to the target domain.

\subsubsection{Neurophysiological interpretability via topographic maps}
\begin{figure*}[!t]
\centering
\subfloat[]{\includegraphics[width=0.45\textwidth]{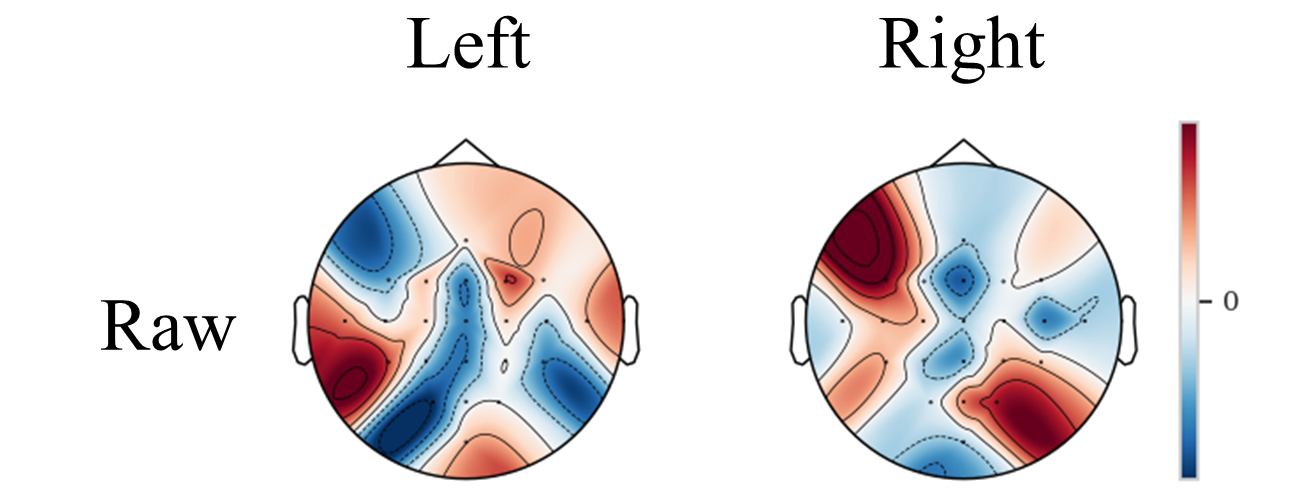}\label{fig:raw_topo}}
\hfill
\subfloat[]{\includegraphics[width=0.45\textwidth]{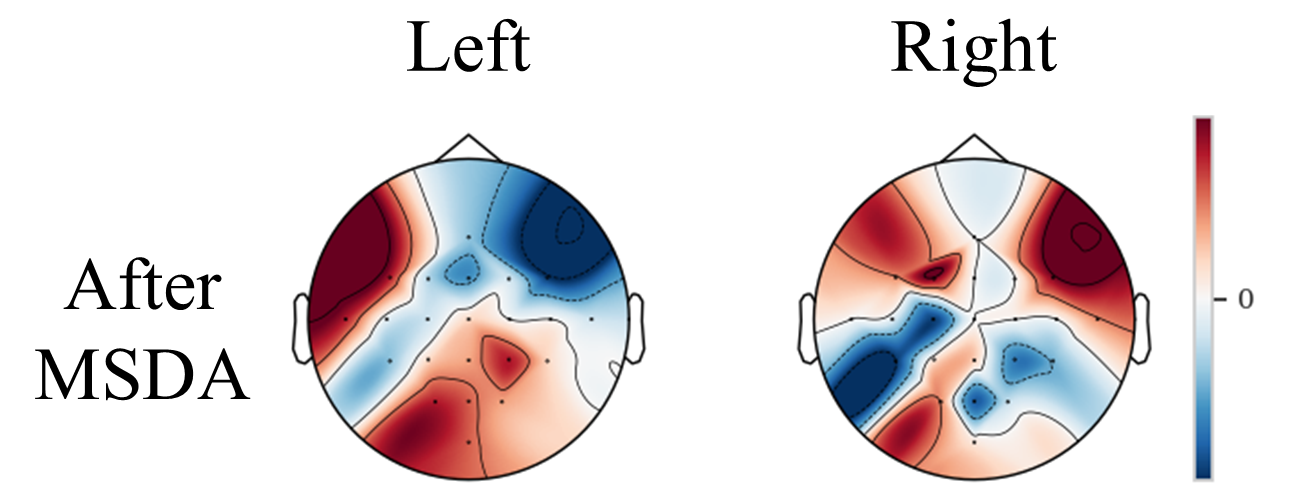}\label{fig:msda_topo}}
\caption{EEG topography for subject 1 from Dataset I with different MI tasks. (a) Raw EEG signals averaged across all trials. (b) Grad-CAM of extracted features after BFM-MSDA.}
\label{fig:topo}
\end{figure*}

To provide neuroscientific insight into the decoding performance, we visualized spatial patterns and model attention on EEG channels. Fig.~\ref{fig:raw_topo} shows the topographic maps of raw EEG data of subject s1 from Dataset I, highlighting sensor-level activity patterns during MI tasks. Fig.~\ref{fig:msda_topo} displays Gradient-weighted Class Activation Mapping (Grad-CAM) \cite{selvarajuGradCAMVisualExplanations2017} results for the features extracted after the proposed BFM-MSDA method. Grad-CAM highlights the EEG channels and temporal regions most influential for model predictions.

Compared to the raw data, our method's Grad-CAM maps exhibit stronger and more focused activations over sensorimotor areas known to be involved in MI, such as the central and precentral regions. In addition, we observe spatial lateralization in S1 between brain hemispheres during left- and right-hand MI, and the class-dependent lateralization is consistent with known MI-related spatial patterns. This demonstrates that the proposed MSDA maintains task-relevant neural signatures rather than learning spurious correlations.

\subsection{Parameter sensitivity analysis}
\label{sec:sensitivity}
A comprehensive sensitivity analysis is conducted to evaluate the impact of the hyperparameters \(\alpha\), \(\beta\), and \(\tau_0\), which balance the FLA loss and the DLA loss, and the point to introduce DLA, respectively.

\begin{figure*}[!t]
\centering
\subfloat[]{\includegraphics[width=0.45\textwidth]{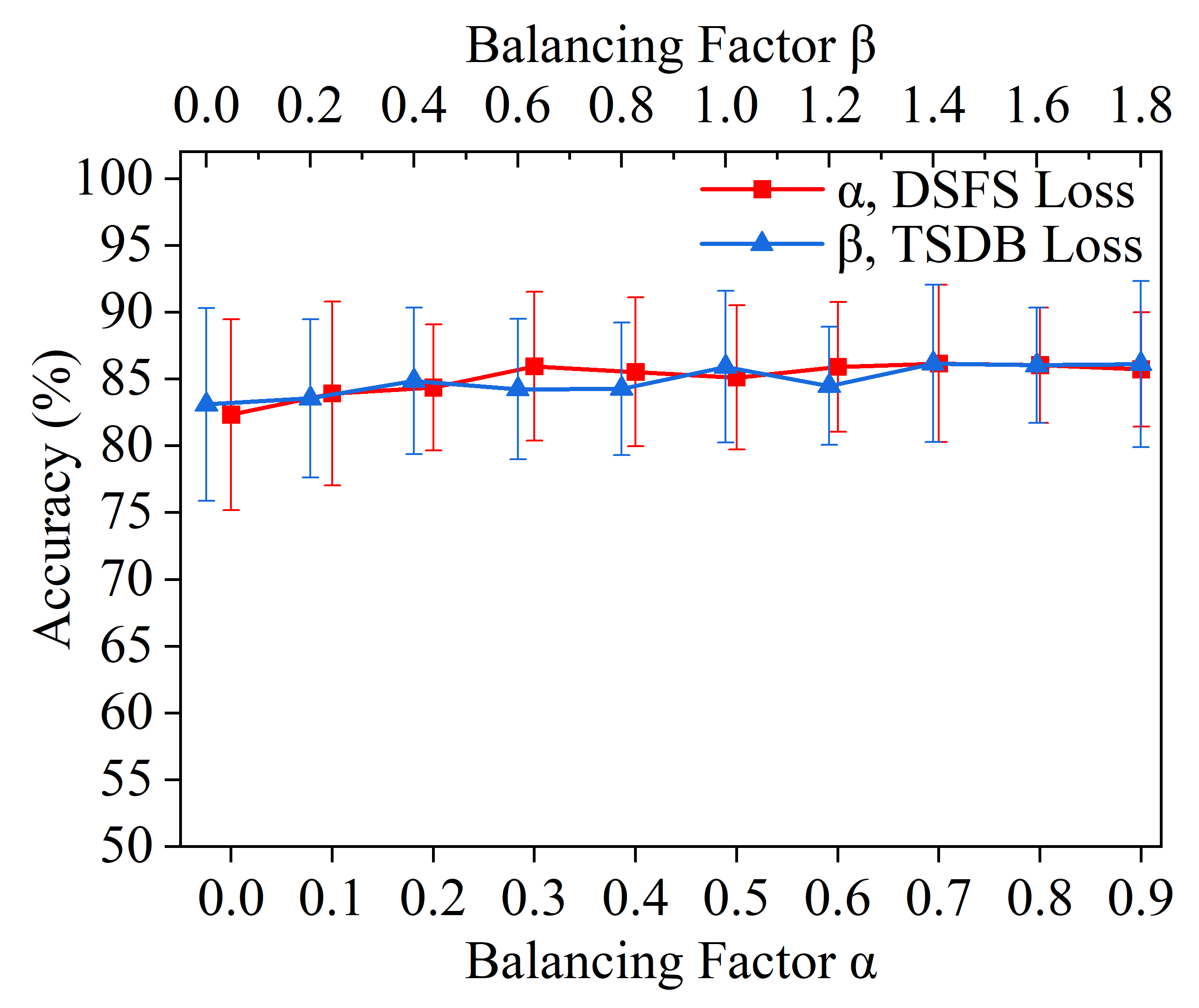}\label{parameter_2a}}
\hfill
\subfloat[]{\includegraphics[width=0.45\textwidth]{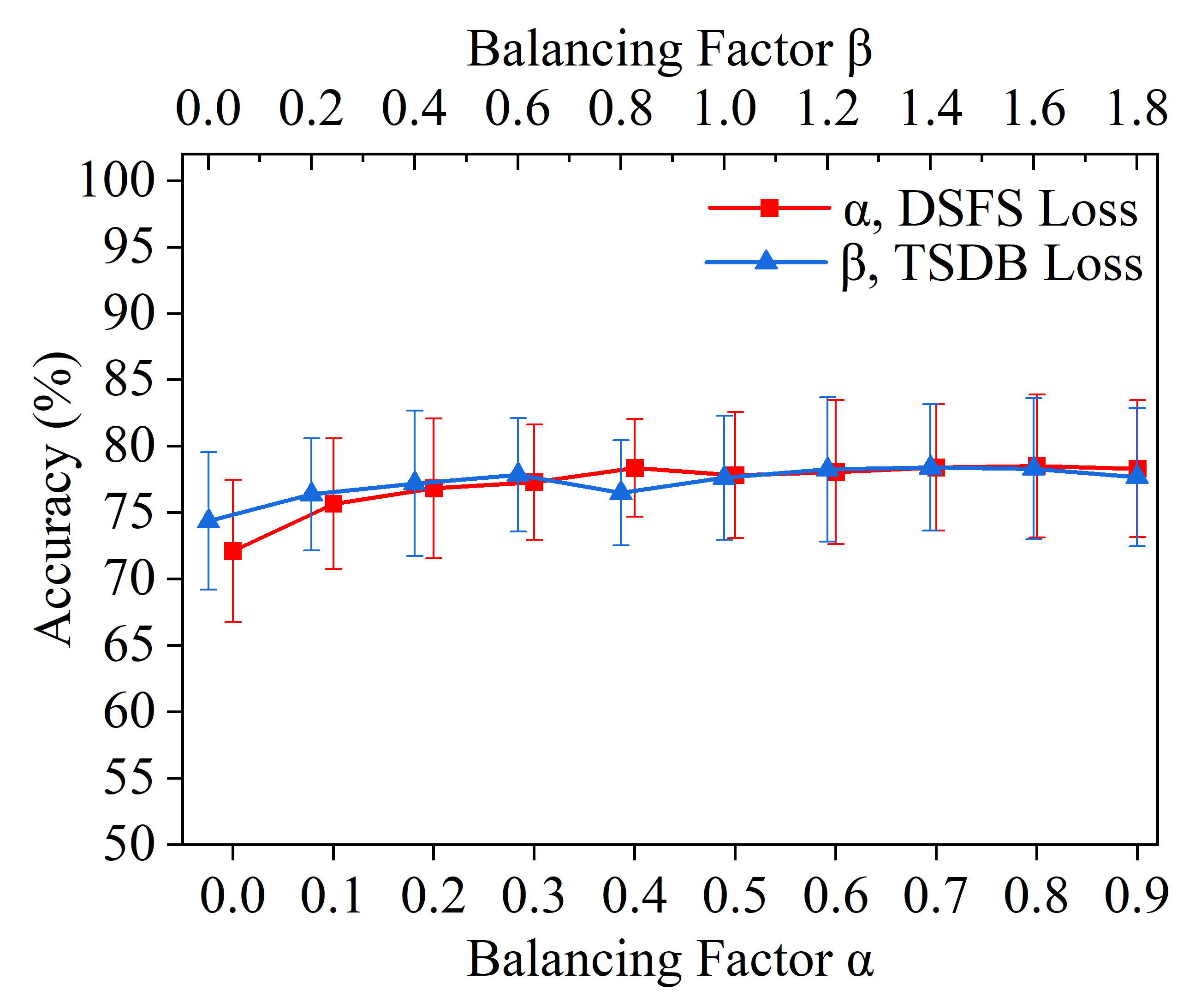}\label{parameter_cho}}\\
\subfloat[]{\includegraphics[width=0.45\textwidth]{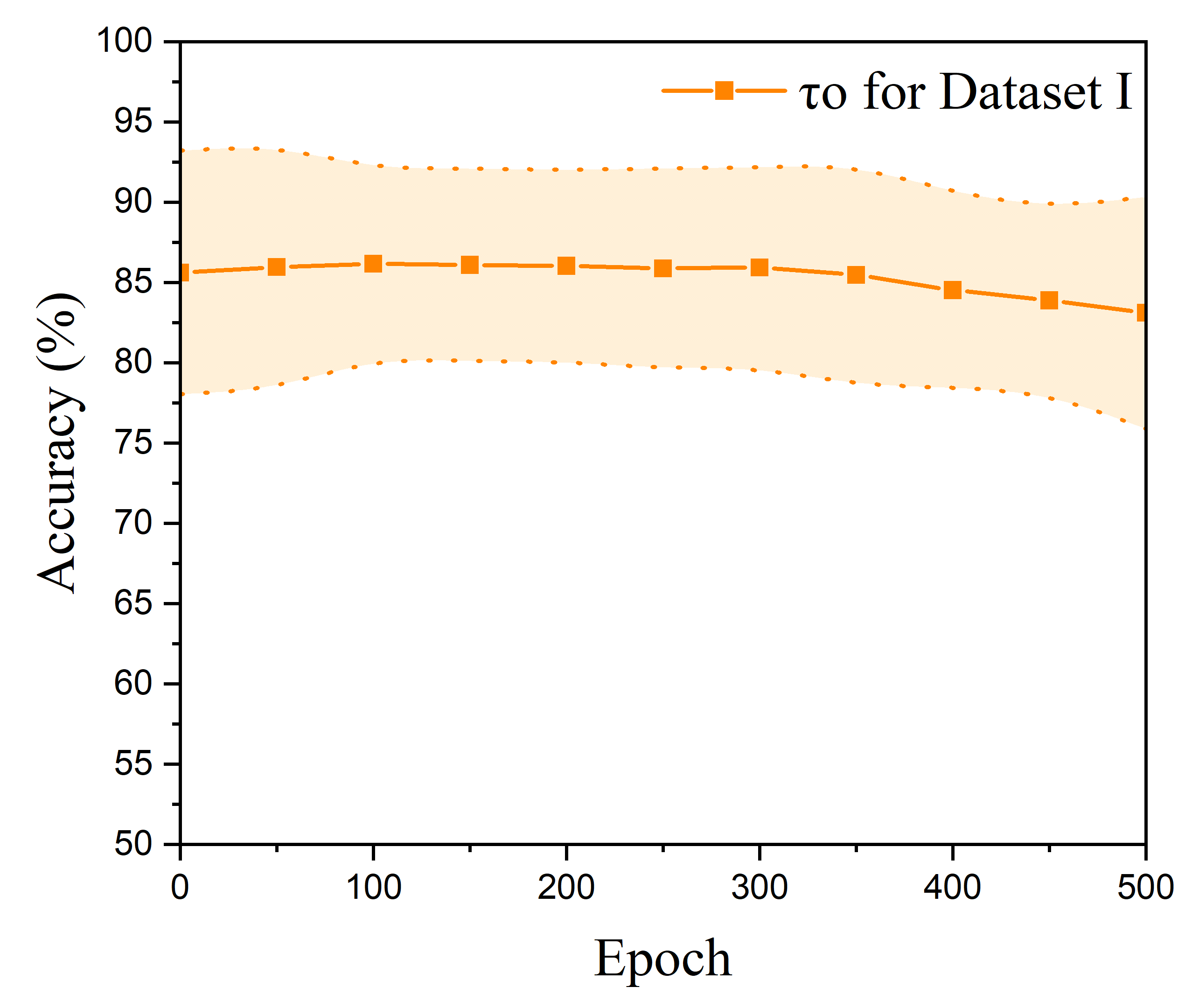}\label{warm-up_I}}
\hfill
\subfloat[]{\includegraphics[width=0.45\textwidth]{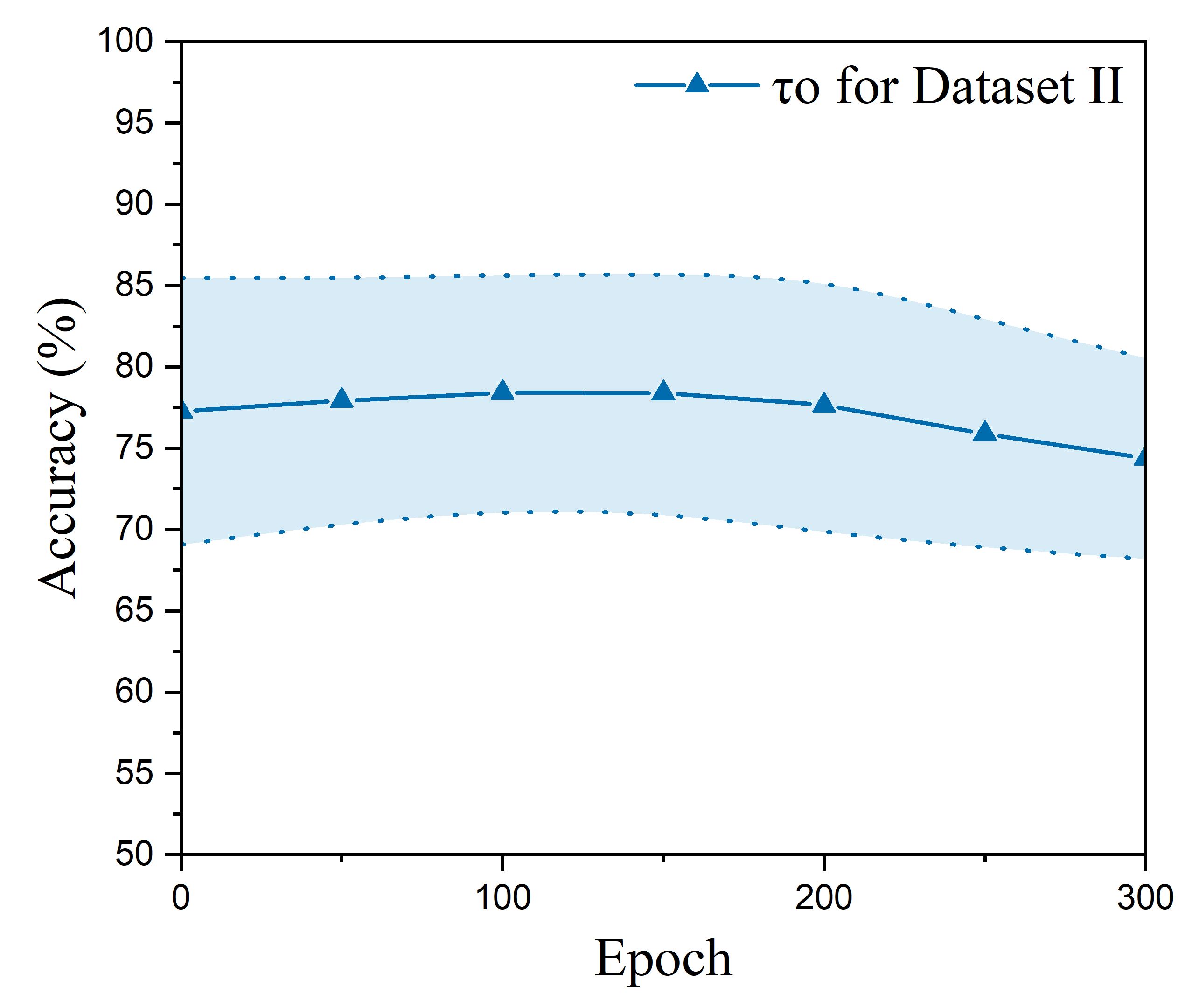}\label{warm-up_II}}
\caption{Parameter sensitivity analysis. (a) Influence of balancing factors for Dataset I. (b) Influence of balancing factors for Dataset II. (c) Influence of transition point \(\tau_0\) for Dataset I. (d) Influence of transition point \(\tau_0\) for Dataset II.}
\label{fig:sensitivity}
\end{figure*}

As shown in Fig.~\ref{parameter_2a} and Fig.~\ref{parameter_cho}, the framework exhibits stable and consistent performance across a wide range of \(\alpha\) and \(\beta\) values once the alignment losses are activated. For both datasets, the lowest classification accuracy occurs when \(\alpha = \beta = 0\), corresponding to the absence of alignment regularization. Introducing the FLA and DLA alignment losses leads to noticeable improvements in accuracy. Consistent with the ablation results in Sec.~\ref{sec:ablation}, the performance gain attributed to the FLA alignment loss is more pronounced than that of the DLA loss on both datasets, though the two are closer in magnitude on Dataset~II. However, after the inclusion of these alignment terms, further tuning of \(\alpha\) and \(\beta\) results in only minor fluctuations in accuracy. Moreover, Figs.~\ref{warm-up_I} and \ref{warm-up_II} indicate stable performances across a reasonable range of \(\tau_0\) values from 50 to 200. There is a good balance for both datasets at \(\tau_0 = 100\), without requiring dataset-specific tuning. This indicates that the proposed framework is robust to the specific choice of these hyperparameters.

\section{Discussion}

\subsection{Effect of BFM-guided source selection}
As shown in Tables~\ref{tab:dataset1} and~\ref{tab:dataset2}, full-backbone fine-tuning of LaBraM and CBraMod produces competitive end-to-end MI decoders, yet the proposed framework still achieves higher average accuracy with a substantially lighter adaptation backbone. These results show that the frozen embedding geometry of a pretrained BFM encodes inter-subject compatibility that can guide source retrieval. Crucially, the pair-level analysis in Sec.~\ref{sec:correlation} establishes this mechanism independently of the full MSDA objective. Across both datasets, sources closer to the target in BFM space achieve significantly higher single-source transferability after target difficulty is controlled (\(\bar{\rho}=-0.304\), \(p=.016\), and \(\bar{\rho}=-0.265\), \(p=.018\), respectively). Hence, BFM distance supplies a principled transferability prior: it retrieves source subjects whose neural representations are more likely to support target decoding before any domain alignment is performed. The ablation results in Sec.~\ref{sec:ablation} translate this pair-level evidence into end-to-end gains: source selection adds 1.04\% on Dataset~I and 2.37\% on Dataset~II, where inter-subject variability is larger. Replacing LaBraM with CBraMod as the frozen selector leaves downstream accuracy in the same range (~\ref{app:cbramod}), indicating that the retrieval principle is not tied to a single BFM architecture.

\subsection{Complementarity of feature-level and decision-level alignment}
The ablation study shows that FLA with CS divergence reduces inter-subject distributional shift, while DLA with conditional CS divergence aligns class-conditional decision structure. FLA provides the larger single-component gain (5.17\% versus 2.83\% on Dataset~I; 4.31\% versus 3.79\% on Dataset~II). Adding DLA on top of FLA improves accuracy by a further 2.02\% on Dataset~I and 3.14\% on Dataset~II, so alignment of \(p(y|\bm{z})\) recovers decision-boundary information that marginal feature alignment does not. The two terms contribute more evenly on Dataset~II, and their combined gain is sub-additive, indicating complementary rather than independent effects. The comparative experiments in Sec.~\ref{sec:kl_comparison} support CS and CCS divergences over KDE-based KL alternatives. The t-SNE and Grad-CAM analyses are consistent with these accuracy gains: target-domain features become more class-discriminative, and model attention remains concentrated over sensorimotor cortex.

\subsection{Scalability to large source repositories}
The large-source-pool experiment demonstrates that increasing the volume of source data does not inherently improve adaptation quality. Instead, as the candidate repository expands, decoding performance increasingly depends on selecting appropriate prior subjects rather than indiscriminately enlarging the source set. As shown in Table~\ref{tab:source_groups} and Fig.~\ref{fig:computationalcost}, accuracy peaks when a compact, target-relevant source set is retained; including too many candidates degrades accuracy and substantially increases training time. From a system-design perspective, constraining the number of transferred source subjects benefits not only classification accuracy but also adaptation efficiency.~\ref{app:complexity} shows that the leading per-epoch cost is $\mathcal{O}(N^2 B^2(d+K))$. On the LOSO, the quadratic saving is $(N/S)^2$ and varies by target.

\subsection{Limitations and future work}
The present findings should be interpreted within the scope of a controlled cross-subject evaluation involving healthy participants and binary left/right MI tasks. Although the proposed framework is expected to be adaptable, generalization to clinical populations such as stroke survivors, where EEG signal characteristics may differ substantially, remains to be seen. Another limitation is that source selection depends on the geometry of the pretrained BFM embedding space. If this space captures nuisance variation, such as recording hardware or montage differences, more strongly than MI-discriminative structure, the selected sources may not be optimal. The pair-level correlations in~\ref{app:correlation} are consistent with this limitation: the association is significant on average but reverses for a minority of targets.

Future work should address the following directions: (i)~evaluating robustness to mismatch across heterogeneous recording setups, (ii)~extending the framework to multi-class MI paradigms and clinical populations, and (iii)~investigating continual source retrieval strategies that accommodate repository growth over time.

\section{Conclusion}

This study presented a BFM-guided source-selective MSDA framework for cross-subject EEG decoding. The proposed BFM-MSDA framework first retrieves target-relevant sources in a pretrained neural representation space and then performs relevance-weighted distribution alignment over the selected sources. By combining CS divergence for marginal feature alignment with conditional CS divergence for decision alignment, the framework addresses both subject-level representation shift and decision-level mismatch.
The experimental results on two public MI-EEG benchmarks show that the proposed framework improves cross-subject decoding over representative baselines. The large-source-pool experiments further demonstrate that poorly matched sources can increase computational cost and degrade transfer. These findings may be useful for other non-stationary pattern-recognition problems where labeled data are distributed across heterogeneous source domains and target labels are unavailable.

\section*{Acknowledgments}
This research was supported by the Innovation Research Group Project of National Natural Science Foundation of China under Grant T2421001 and the Key Project of the Chongqing Technology Innovation and Application Development Special Program under Grant CSTB2025TIAD-JM011. At the same time, the authors gratefully acknowledge financial support from China Scholarship Council, China.

\appendix

\makeatletter
\renewcommand{\p@section}{Appendix~}
\renewcommand{\p@subsection}{Appendix~}
\makeatother
\counterwithin*{equation}{section}
\counterwithin*{figure}{section}
\counterwithin*{table}{section}
\renewcommand{\theequation}{\thesection.\arabic{equation}}
\renewcommand{\thefigure}{\thesection.\arabic{figure}}
\renewcommand{\thetable}{\thesection.\arabic{table}}

\section{Estimator of Cauchy-Schwarz and Conditional Cauchy-Schwarz Divergences}
\label{app:ccs}
In practice, the probability density function \(p(y|\mathbf{z})\) is unknown and must be estimated from finite samples. We adopt kernel-based empirical estimators for CS and CCS divergences \cite{yuConditionalCauchySchwarzDivergence2024}, enabling efficient and nonparametric computation.

\subsection{Empirical Estimator of CS Divergence}
Given extracted features from two domains, \(\{\mathbf{z}_i^s\}_{i=1}^{B_s}\) from the source and \(\{\mathbf{z}_j^t\}_{j=1}^{B_t}\) from the target, where \(B_s\) and \(B_t\) denote the source and target mini-batch sizes, the CS divergence can be empirically estimated as:
\begin{equation} \label{eq:empirical_cs}
\begin{aligned}
&\widehat{D}_{\mathrm{CS}}(p^s(\mathbf{z}); p^t(\mathbf{z}))
= \log \left(\frac{1}{B_s^2} \sum_{i=1}^{B_s} \sum_{j=1}^{B_s} \kappa(\mathbf{z}_i^s, \mathbf{z}_j^s)\right) \\
&\quad + \log \left(\frac{1}{B_t^2} \sum_{i=1}^{B_t} \sum_{j=1}^{B_t} \kappa(\mathbf{z}_i^t, \mathbf{z}_j^t)\right)\\
&\quad - 2 \log \left(\frac{1}{B_s B_t} \sum_{i=1}^{B_s} \sum_{j=1}^{B_t} \kappa(\mathbf{z}_i^s, \mathbf{z}_j^t)\right),
\end{aligned}
\end{equation}
where \(\kappa(\cdot, \cdot)\) is a Gaussian kernel
\[
\kappa_{\sigma}(\mathbf{z}, \mathbf{z}') = \exp\left(-\frac{\|\mathbf{z} - \mathbf{z}'\|_2^2}{2\sigma^2}\right),
\]
and the bandwidth parameter \(\sigma\) for all Gaussian kernels \(\kappa(\cdot, \cdot)\) is determined using the median heuristic: \(\sigma = \text{median}(\{\|\mathbf{z}_i - \mathbf{z}_j\|_2 : i \neq j\})\), where the median is computed over all pairwise distances within each mini-batch.

\subsection{Empirical Estimator of CCS Divergence}
Given features and predicted outputs from source and target domains, \(\{(\mathbf{z}_i^s, \hat{y}_i^s)\}_{i=1}^{B_s}\) and \(\{(\mathbf{z}_j^t, \hat{y}_j^t)\}_{j=1}^{B_t}\), define Gram matrices as:
\begin{align}
K^s_{ij} &= \kappa(\mathbf{z}_i^s, \mathbf{z}_j^s),\quad
L^s_{ij} = \ell(\hat{y}_i^s, \hat{y}_j^s), \notag \\
K^t_{ij} &= \kappa(\mathbf{z}_i^t, \mathbf{z}_j^t),\quad
L^t_{ij} = \ell(\hat{y}_i^t, \hat{y}_j^t), \notag \\
K^{st}_{ij} &= \kappa(\mathbf{z}_i^s, \mathbf{z}_j^t),\quad
L^{st}_{ij} = \ell(\hat{y}_i^s, \hat{y}_j^t), \notag \\
K^{ts}_{ij} &= \kappa(\mathbf{z}_i^t, \mathbf{z}_j^s),\quad
L^{ts}_{ij} = \ell(\hat{y}_i^t, \hat{y}_j^s), \notag
\end{align}
where \(\kappa(\cdot, \cdot)\) and \(\ell(\cdot, \cdot)\) are kernel functions for features and predicted outputs, respectively.

The empirical CCS divergence estimator is approximated by:
\begin{equation} \label{eq:empirical_ccs}
\begin{aligned}
&\widehat{D}_{\mathrm{CCS}}(p^s(\hat{y}|\mathbf{z}); p^t(\hat{y}|\mathbf{z})) \approx \log \left( \sum_{j=1}^{B_s} \frac{\sum_{i=1}^{B_s} K^s_{ji} L^s_{ji}}{\left(\sum_{i=1}^{B_s} K^s_{ji}\right)^2} \right) \\
&\quad + \log \left( \sum_{j=1}^{B_t} \frac{\sum_{i=1}^{B_t} K^t_{ji} L^t_{ji}}{\left(\sum_{i=1}^{B_t} K^t_{ji}\right)^2} \right) \\
&\quad - \log \left( \sum_{j=1}^{B_s} \frac{\sum_{i=1}^{B_t} K^{st}_{ji} L^{st}_{ji}}{\left(\sum_{i=1}^{B_s} K^s_{ji}\right) \left(\sum_{i=1}^{B_t} K^{st}_{ji}\right)} \right) \\
&\quad - \log \left( \sum_{j=1}^{B_t} \frac{\sum_{i=1}^{B_s} K^{ts}_{ji} L^{ts}_{ji}}{\left(\sum_{i=1}^{B_s} K^{ts}_{ji}\right) \left(\sum_{i=1}^{B_t} K^t_{ji}\right)} \right).
\end{aligned}
\end{equation}

\section{Kernel Density Estimation-based Estimator of Joint KL Divergence}
\label{app:kde_kl}
KL divergence can also be estimated in a nonparametric manner by the kernel density estimation (KDE) approach.

\subsection{Joint KDE Formulation}
\label{app:joint_kde}
Given the source domain \(D_s=\{(\bm{z}_i^s,\hat{\bm{y}}_i^s)\}_{i=1}^{M_s}\) and the target domain \(D_t=\{(\bm{z}_j^t,\hat{\bm{y}}_j^t)\}_{j=1}^{M_t}\), denote latent features as \(\bm{z}=f(\bm{x})\in\mathbb{R}^d\) and predicted probability distributions as \(\hat{\bm{y}}=g(\bm{z})\) with \(K\) classes. The joint densities are estimated by:
\begin{equation}
\begin{aligned}
\hat p_s(\bm{z},\hat{\bm{y}}) &=
\frac{1}{M_s}\sum_{i=1}^{M_s}
\mathcal{K}_{z,h_{z,i}^s}\!\left(\bm{z}-\bm{z}_i^s\right) \\
&\qquad \times \mathcal{K}_{y,h_{y,i}^s}\!\left(\bm{y}-\hat{\bm{y}}_i^s\right), \\
\hat q_t(\bm{z},\hat{\bm{y}}) &=
\frac{1}{M_t}\sum_{j=1}^{M_t}
\mathcal{K}_{z,h_{z,j}^t}\!\left(\bm{z}-\bm{z}_j^t\right) \\
&\qquad \times \mathcal{K}_{y,h_{y,j}^t}\!\left(\bm{y}-\hat{\bm{y}}_j^t\right),
\end{aligned}
\label{eq:joint_kde_separable}
\end{equation}
where \(\mathcal{K}(\cdot)\) is a Gaussian kernel in \(\mathbb{R}^{d+K}\) with adaptive bandwidth determined by the median heuristic. This separable form is equivalent to applying a Gaussian kernel over the concatenated vector \(\bm{u} = [\bm{z};\hat{\bm{y}}]\in\mathbb{R}^{d+K}\), while allowing distinct bandwidths \(h_z\) and \(h_y\).

\subsection{Joint KL Divergence Estimator}
\label{app:kl_estimator}
We denote each concatenated feature--label sample as \(\bm{u}=[\bm{z};\hat{\bm{y}}]\), and evaluate the estimated densities \(\hat{p}_s(\bm{u})\) and \(\hat{q}_t(\bm{u})\) at these joint points. With the KDE estimates in Eq.~\eqref{eq:joint_kde_separable}, the forward KL divergence is estimated as:
\begin{equation}
\widehat{D}_{\mathrm{KL}}
\!\left(\hat{p_s}(\bm{u})\,\Vert\,\hat{q_t}(\bm{u})\right)
=
\frac{1}{|\mathcal{B}_s|}
\sum_{\bm{u}\in\mathcal{B}_s}
\log
\frac{\hat p_s(\bm{u})}{\hat q_t(\bm{u})}.
\label{eq:joint_kl_forward}
\end{equation}
where \(\mathcal{B}_s\) is a mini-batch of source samples. The same batch-wise KDE evaluation policy as the CS estimator is adopted.

\section{Robustness Analyses of BFM-Based Retrieval}
\label{app:correlation}

The primary within-target analysis and its empirical support for BFM-based retrieval are reported in Sec.~\ref{sec:correlation}. This appendix documents its experimental protocol and statistical test, then provides complementary analyses using alternative pair groupings and characterizes variation in the relationship across targets.

\subsection{Protocol and statistical testing}

For every ordered source--target pair, the backbone is trained on labelled data from the source subject alone for 200 epochs and evaluated on the target without domain adaptation. Each experiment is repeated over five random seeds. Transferability is the mean accuracy over the last ten epochs, averaged across seeds. This produces \(9 \times 8=72\) pairs on Dataset~I and \(10 \times 9=90\) pairs on Dataset~II.

To control target difficulty, Spearman's rank correlation between BFM distance and transferability is calculated separately for each target and then averaged across targets. Statistical significance is evaluated by independently shuffling transferability values within each target for 20,000 permutations. This blocked permutation design preserves the source--target structure while testing whether the observed mean within-target correlation differs from zero.

\begin{table}[t]
  \centering
  \small
  \caption{Supplementary rank correlations between BFM distance and single-source transferability under alternative pair groupings. Negative values indicate that nearer sources transfer better.}
  \label{tab:corr}
  \begin{tabular}{llrrr}
    \toprule
    Dataset & Analysis & $\bar{\rho}$ & $p$ & Blocks \\
    \midrule
    \multirow{2}{*}{I}
      & Within source   & $-0.283$ & $0.026$ & 9 \\
      & Pooled (72 pairs) & $-0.350$ & $0.002$ & -- \\
    \midrule
    \multirow{2}{*}{II}
      & Within source   & $-0.132$ & $0.250$ & 10 \\
      & Pooled (90 pairs) & $-0.182$ & $0.088$ & -- \\
    \bottomrule
  \end{tabular}
\end{table}

Reversing the grouping provides a consistency check. Holding the \emph{source} fixed and ranking its candidate targets gives $\bar{\rho} = -0.283$ ($p = .026$) on Dataset~I, close to the within-target value; the conclusion therefore does not depend on which side of the pair is held constant. On Dataset~II the within-source correlation is weaker and not significant ($\bar{\rho} = -0.132$, $p = .250$), which is expected: with 52 candidate subjects reduced to a 10-subject subset, the retained targets span a narrower range of distances than the sources do.

\subsection{Effect Size and Heterogeneity}

The per-target relationship is heterogeneous: correlations range from $-0.95$ to $+0.24$ on Dataset~I (6 of 9 targets negative) and from $-0.72$ to $+0.87$ on Dataset~II (7 of 10 targets negative). This variation motivates the adaptive soft-gating scheme, which uses BFM distance as a target-aware prior while allowing the downstream dual-alignment model to integrate information from multiple retrieved sources.




\section{Effect of the Number of Selected Sources}
\label{app:k_curves}

The correlation above establishes that the BFM distance ranks sources. This section measures what that ranking is worth when the number of admitted sources is varied. It expands the unadapted comparison cited after the ablation in Sec.~\ref{sec:ablation}.

\subsection{Protocol}

This experiment retains the unadapted single-source setting of Equation~\eqref{eq:transferability} and extends it to source sets. For each target we order the candidate sources by BFM distance, train on the nearest $k$ for $k = 1 \dots n$, and compare against training on $k$ sources drawn at random (12 draws per $k$, averaged). Sources are pooled and trained with cross-entropy alone; neither alignment term nor the adaptive source weighting of the full method is applied. Both arms use the identical recipe and the same 200-epoch schedule as in ~\ref{app:correlation}, so the curves are directly comparable to each other and must coincide at $k = n$, where every ordering holds the same set.

Isolating selection from adaptation is deliberate: it measures what the BFM ordering contributes on its own, before the alignment terms have a chance to compensate for a poorly chosen source set. The numbers here are therefore \emph{not} interchangeable with the ablation in Sec.~\ref{sec:ablation}, which compares selection against no selection inside the complete BFM-MSDA model.

\subsection{Results}

\begin{figure}[t]
  \centering
  \includegraphics[width=\columnwidth]{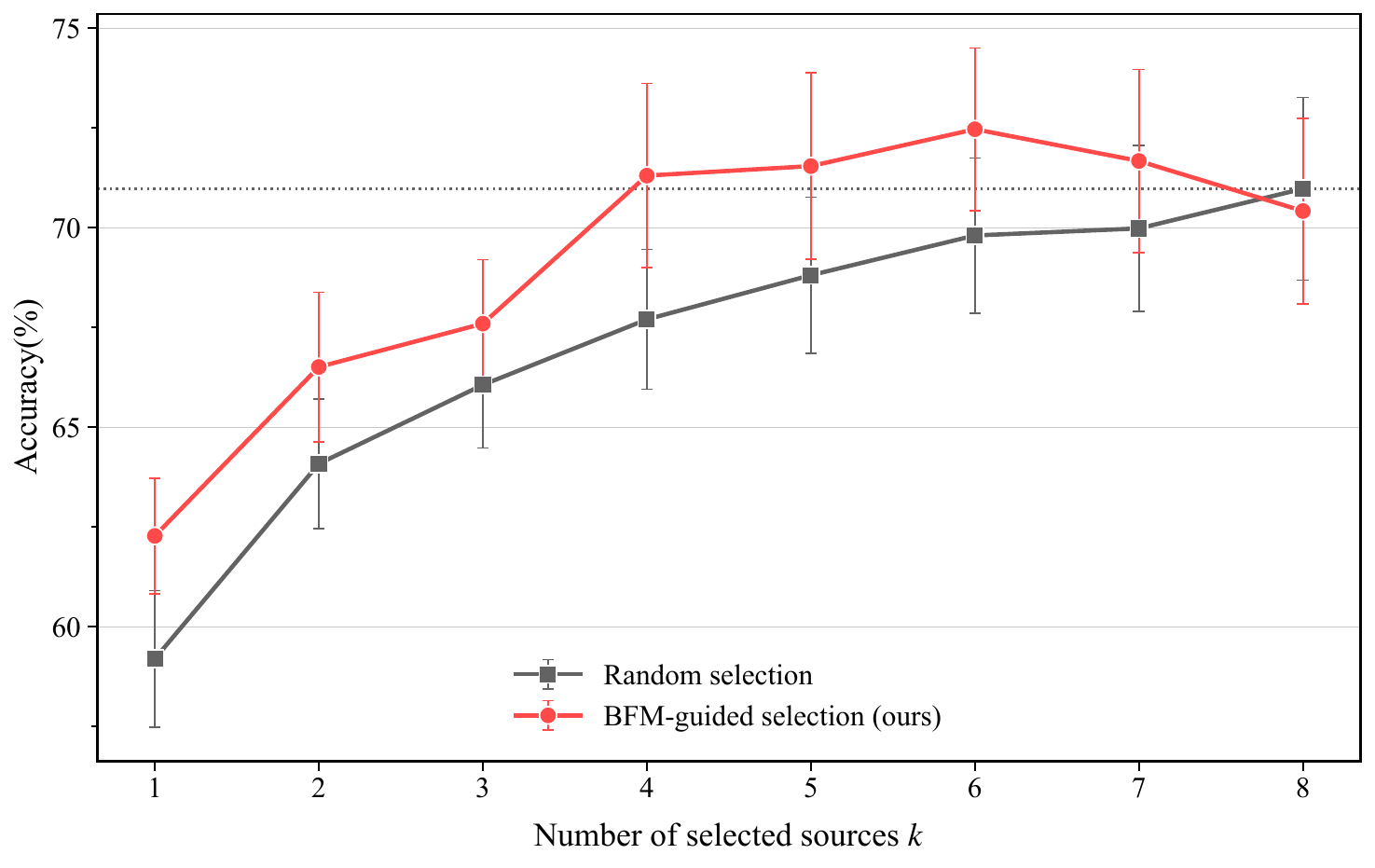}
  \caption{Target accuracy against the number of selected sources on Dataset~I, without domain adaptation. BFM-guided selection is above random selection at every $k$ below the full pool and peaks at $k = 6$. The dotted line marks the best accuracy random selection attains at any $k$. Error bars are the standard error across targets.}
  \label{fig:acc_vs_k_2a}
\end{figure}

\begin{figure}[t]
  \centering
  \includegraphics[width=\columnwidth]{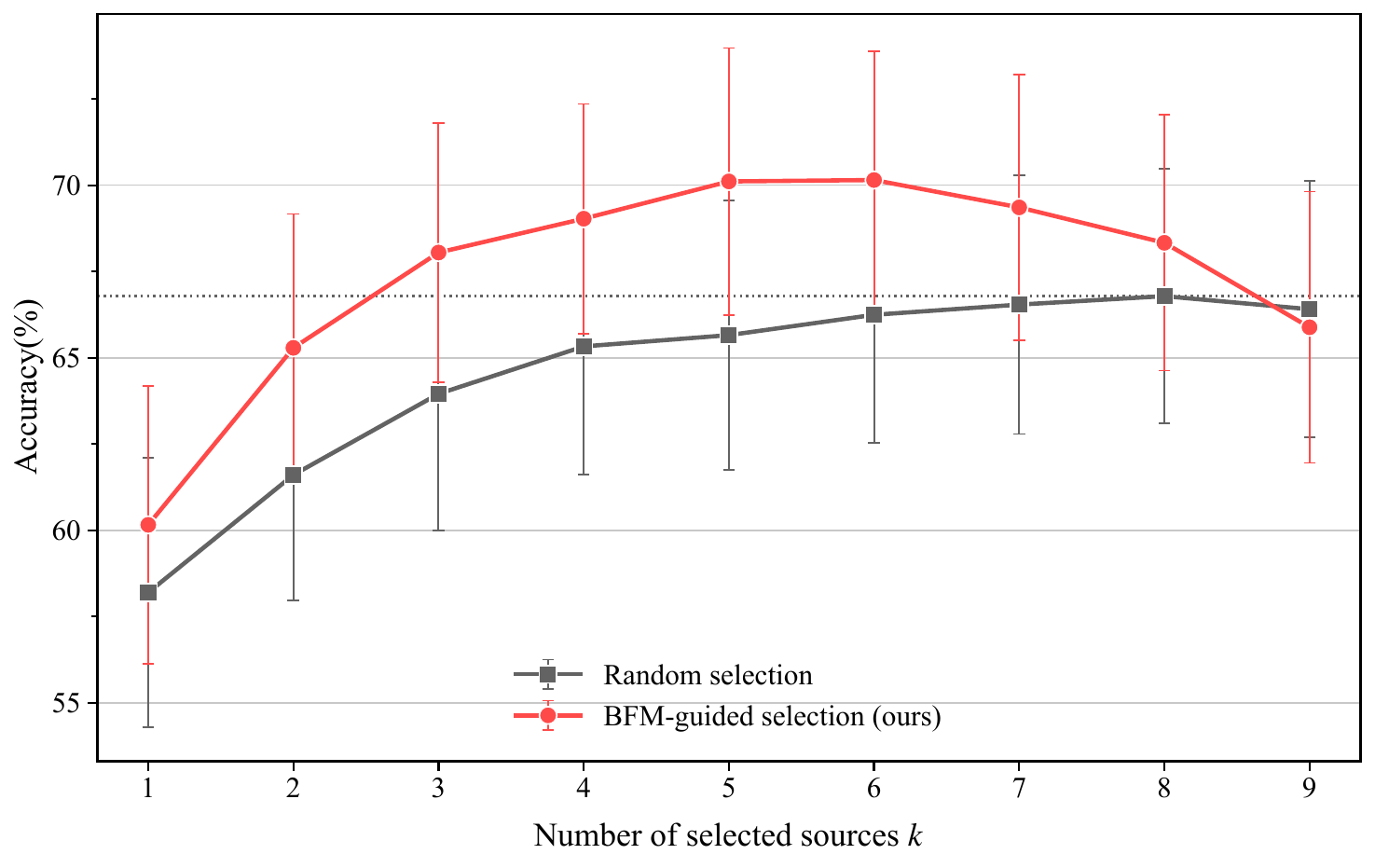}
  \caption{Target accuracy against the number of selected sources on Dataset~II, without domain adaptation. Accuracy falls $4.3$ points from its maximum at $k = 6$ to $k = 9$. Error bars are the standard error across targets.}
  \label{fig:acc_vs_k_cho}
\end{figure}

Figures~\ref{fig:acc_vs_k_2a} and~\ref{fig:acc_vs_k_cho} show three effects common to both datasets. First, BFM-guided selection is above random selection at every $k$ below the full pool, by $1.5$--$3.6$ points on Dataset~I and $1.5$--$4.5$ points on Dataset~II. Second, both curves rise and then fall: additional sources supply more training data up to a point, beyond which the sources being admitted are distant enough that negative transfer outweighs the gain. Accuracy peaks at $k = 6$ on both datasets, and using the entire pool costs $2.0$ points on Dataset~I and $4.3$ points on Dataset~II relative to those peaks. Third, the curves converge at $k = n$ to within $0.6$ points, confirming that the two arms differ only in the order in which sources are admitted.

Selection also reaches a given accuracy with fewer sources. The best accuracy random selection attains at any $k$ is matched by BFM-guided selection at $k = 4$ on Dataset~I and $k = 3$ on Dataset~II, against $k = 8$ for random selection in both cases. Since training cost scales with the number of sources in the batch, this halves or better the cost of reaching that accuracy.

These $k$ values are diagnostic of the unadapted pooling regime. They are not a recommendation to replace the adaptive soft-gating rule of the full method, which additionally uses relevance-weighted CS/CCS alignment.

\paragraph{Relation to the ablation in Sec.~\ref{sec:ablation}.}
The ablation attributes $+1.04$ points on Dataset~I and $+2.37$ on Dataset~II to source selection within the complete model. The values here are larger --- $2.0$ and $4.3$ points --- because the alignment terms are switched off. The adaptive weighting $\mathrm{softmax}(-D_{\mathrm{CS}}(p_s(\bm{z})\|p_t(\bm{z})))$ already suppresses distant sources during training, so a poor source set costs the full model less than it costs the unadapted baseline. The two measurements agree on direction, on relative magnitude between the datasets (Dataset~II benefits roughly twice as much in both), and on the mechanism; they differ in scale by the amount the alignment recovers. Read together they indicate that selection and alignment address the same failure mode at different stages, and that selection matters most where alignment is absent or the candidate pool is large.
\section{Complexity of Pairwise Alignment}
\label{app:complexity}

Let $N$ be the number of \emph{selected} sources, $S$ the number of available candidates, $B$ the mini-batch size, $d$ the feature dimension, and $K$ the number of classes.

The per-epoch cost of BFM-MSDA is dominated by pairwise CS and CCS divergences. A kernel CS estimate between two batches of size $B$ in dimension $d$ costs $\mathcal{O}(B^2 d)$.
\begin{itemize}
    \item \textbf{Feature-level alignment (FLA):} source--target pairs cost $\mathcal{O}(N B^2 d)$ and source--source pairs cost $\mathcal{O}(N^2 B^2 d)$.
    \item \textbf{Decision-level alignment (DLA):} the same pairing in logit space costs $\mathcal{O}(N B^2 K + N^2 B^2 K)$.
    \item \textbf{Classification loss:} $\mathcal{O}(N B d)$.
\end{itemize}
The leading term is therefore
\[
\mathcal{O}\bigl(N^2 B^2 (d+K) + N B d\bigr).
\]
With $B$, $d$, and $K$ fixed by the experimental setup, training cost is quadratic in the number of \emph{selected} sources $N$.

Without source selection, $N=S$ and the cost is $\mathcal{O}(S^2 B^2(d+K))$. The reduction factor depends on the selection rule and on the size of the candidate pool:
\begin{itemize}
    \item On the 42-subject GigaDB repository, the controlled study in Sec.~\ref{sec:large_pool} uses a fixed 25th-percentile cutoff, so $N \approx 0.25S$ and
    \[
    \frac{(0.25S)^2}{S^2} = \frac{1}{16}.
    \]
    \item On the LOSO splits of Datasets~I and~II, $N$ is set by adaptive soft-gating and varies across targets; the quadratic saving is then $(N/S)^2$ for that target, not a universal $1/16$.
\end{itemize}
This scaling is consistent with the empirical timing in Fig.~\ref{fig:computationalcost}: training time per epoch rises super-linearly as more distant sources are admitted.

\section{Cross-BFM Source Selection with CBraMod}
\label{app:cbramod}

To test whether BFM-guided selection is tied to a single foundation model, we keep the downstream MSDA architecture and training recipe unchanged and replace only the embedding used to compute source--target distances. LaBraM is swapped for CBraMod; selected sources are then fed to the same EEGNet-based BFM-MSDA objective. This CBraMod-as-selector experiment is distinct from the full-backbone CBraMod decoder reported in Tables~\ref{tab:dataset1} and~\ref{tab:dataset2}.

CBraMod uses criss-cross attention to model spatial and temporal EEG structure separately, whereas LaBraM treats channel patches with standard self-attention. Both models expect 200\,Hz input and 200-sample embeddings, so the same preprocessing is applied.

\subsection{Pipeline on Dataset~I}

\begin{figure*}[!t]
\centering
\includegraphics[width=0.8\textwidth]{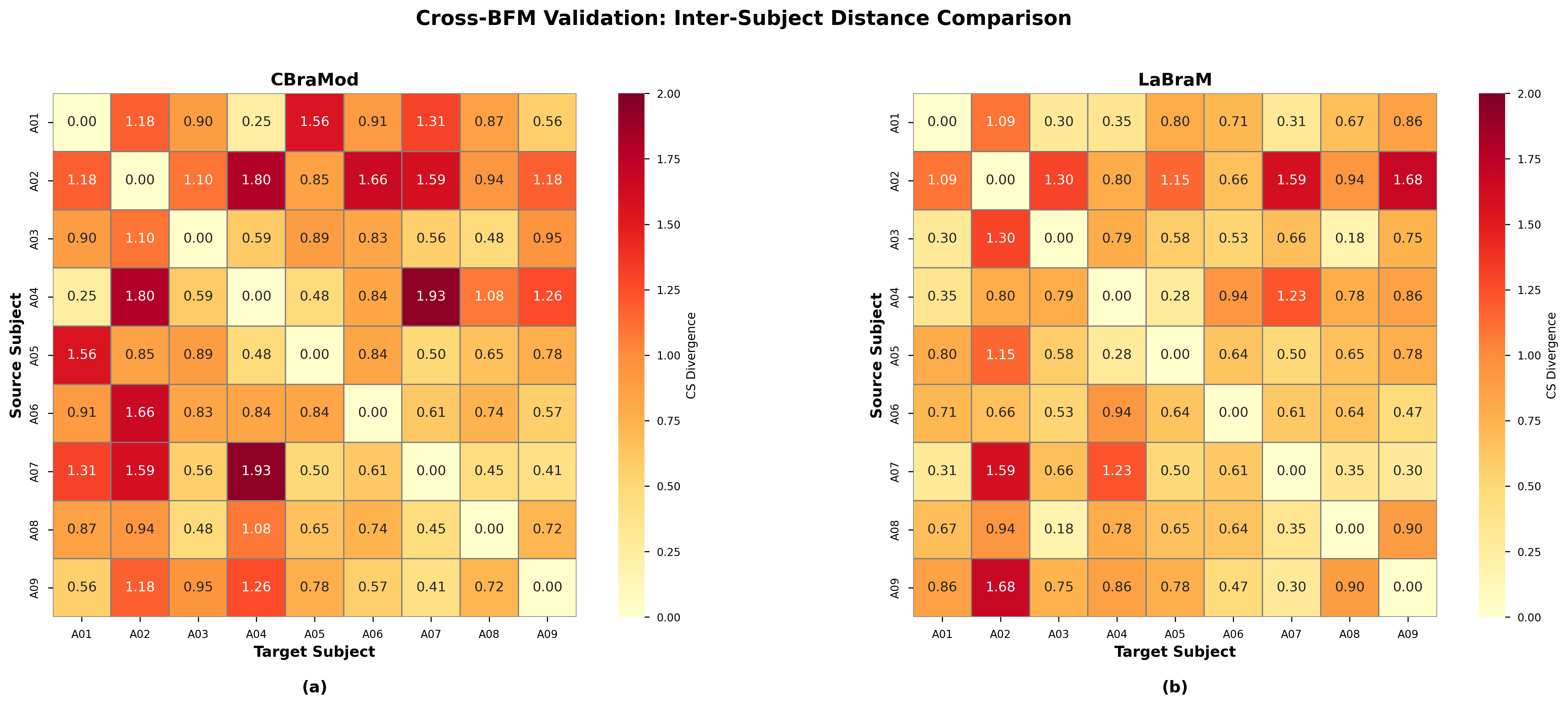}
\caption{Inter-subject CS distances on Dataset~I. (a)~CBraMod embeddings. (b)~LaBraM embeddings.}
\label{fig:cross_bfm_heatmaps}
\end{figure*}

\begin{description}
    \item[Preprocessing.] Raw 22-channel EEG is bandpass-filtered, epoched, and resampled to 200\,Hz as in Sec.~\ref{subsec:source_selection}. Trials are reshaped to $22\ \mathrm{channels} \times 3\ \mathrm{segments} \times 200\ \mathrm{samples}$ to match CBraMod's input layout.

    \item[Pooling.] Pretrained CBraMod weights are used with the final projection removed. Each trial is passed through the 12-layer criss-cross encoder. Hierarchical mean pooling over the spatial and temporal axes produces one 200-dimensional trial-level vector. As with LaBraM, these trial-level vectors are retained as a set and used to estimate $p(\mathbf{h})$.

    \item[Distance.] Pairwise CS divergences use the same kernel estimator as the LaBraM experiments. The two inter-subject distance matrices are shown in Fig.~\ref{fig:cross_bfm_heatmaps}.

    \item[Source selection.] Both embeddings use the same adaptive soft-gating rule as Sec.~\ref{subsec:source_selection}. Distances $d_{s,t}$ are converted to relevance weights by Eq.~\eqref{eq:relevanceweight}, with temperature $\tau_t$ equal to the median source--target distance, and source $s$ is kept if $\tilde{w}_{s,t}>1/S$. If the rule returns an empty set, the top-3 sources are used. $N$ is therefore target-dependent.
\end{description}
\begin{table*}[!t]
\centering
\caption{Adaptive soft-gating source sets and downstream BFM-MSDA accuracies on Dataset~I when source--target distances come from LaBraM or from CBraMod. Overlap is the Jaccard index. LaBraM accuracies match Table~\ref{tab:dataset1}.}
\label{tab:cross_bfm_validation}
\begingroup\small\setlength{\tabcolsep}{4pt}\renewcommand{\arraystretch}{1.0}%
\begin{tabular}{@{}lccccc@{}}
\toprule
\textbf{Target} & \multicolumn{2}{c}{\textbf{LaBraM-based selection}} & \multicolumn{2}{c}{\textbf{CBraMod-based selection}} & \textbf{Overlap} \\
\cmidrule(lr){2-3} \cmidrule(lr){4-5}
& \textbf{Selected sources} & \textbf{Acc.\ (\%)} & \textbf{Selected sources} & \textbf{Acc.\ (\%)} & \textbf{(\%)} \\
\midrule
S1 & S3, S4, S7 & 83.19 & S3, S4, S8 & 84.27 & 50.0 \\
S2 & S1, S4, S6, S8 & 76.53 & S1, S3, S5, S8 & 74.32 & 33.3 \\
S3 & S2, S6, S8 & 91.74 & S4, S6, S8 & 89.08 & 50.0 \\
S4 & S1, S5, S8 & 83.26 & S1, S3, S5, S6 & 82.15 & 40.0 \\
S5 & S3, S4, S7 & 89.37 & S4, S7, S9 & 86.94 & 50.0 \\
S6 & S3, S5, S7 & 81.46 & S3, S7, S8 & 80.35 & 50.0 \\
S7 & S1, S5, S8, S9 & 92.50 & S5, S8, S9 & 88.82 & 75.0 \\
S8 & S3, S7 & 90.14 & S3, S7, S9 & 90.53 & 66.7 \\
S9 & S3, S6, S7 & 87.29 & S1, S6, S7, S8 & 86.64 & 40.0 \\
\midrule
\textbf{Average} & --- & \textbf{86.16} & --- & \textbf{84.79} & \textbf{50.6} \\
\midrule
\multicolumn{6}{l}{Distance correlation \quad Pearson $r=0.546$,\ Spearman $\rho=0.594$} \\
\bottomrule
\end{tabular}
\endgroup
\end{table*}

The two embeddings agree only moderately ($r=0.546$, $\rho=0.594$). Downstream accuracy remains in the same range ($86.16\%$ with LaBraM selection versus $84.79\%$ with CBraMod selection), so the MSDA stage is not brittle to the choice of BFM. The $1.37$ point gap indicates that LaBraM distances are the stronger selector on this split, which is why they remain the default in the main experiments.
\subsection{Overlap and downstream accuracy}

Let $A_t$ and $B_t$ be the source sets selected for target $t$ by LaBraM and CBraMod. Overlap is the Jaccard index
\begin{equation}
J(A_t,B_t) = \frac{|A_t \cap B_t|}{|A_t \cup B_t|},
\end{equation}
which penalizes both extra and missing sources. We also report Pearson and Spearman correlations between the two distance matrices over all LOSO pairs $(s,t)$.

\bibliographystyle{elsarticle-num}

\bibliography{cas-refs}

\end{document}